%% file: main.tex
\documentclass{article}

\usepackage[utf8]{inputenc}
\usepackage[T1]{fontenc}
\usepackage{hyperref}

\usepackage{amsfonts}       % blackboard math symbols
\usepackage{lmodern}
\usepackage{amsmath}
\usepackage{amssymb}
\usepackage{amsthm}
\usepackage{nicefrac}       % compact symbols for 1/2, etc.
\usepackage{changepage}
\usepackage{rotating}

\usepackage{microtype}
\usepackage{graphicx}
\usepackage{booktabs} % for professional tables

\usepackage{xspace}
\usepackage{wrapfig}
\usepackage{multirow}
\usepackage{caption, subcaption}
\usepackage[dvipsnames]{xcolor}
\usepackage{colortbl}
\usepackage{enumitem}

\usepackage{pifont}% http://ctan.org/pkg/pifont

\newcommand{\method}{\texttt{GPSE}\xspace}
\newcommand{\Rsq}{R$^2$\xspace}

\definecolor{dark2green}{RGB}{27, 158, 119}
\definecolor{dark2orange}{RGB}{225, 95, 2}
\definecolor{dark2purple}{rgb}{0.4, 0.4, 0.8}
\definecolor{dark2pink}{RGB}{212, 17, 89}

\usepackage[accepted]{icml2024}

\usepackage{amsmath}
\usepackage{amssymb}
\usepackage{mathtools}
\usepackage{amsthm}

\usepackage[capitalize,noabbrev]{cleveref}

\theoremstyle{plain}
\newtheorem{theorem}{Theorem}[section]
\newtheorem{proposition}[theorem]{Proposition}

\theoremstyle{definition}
\newtheorem{definition}[theorem]{Definition}

\theoremstyle{remark}

\usepackage[textsize=tiny]{todonotes}

\icmltitlerunning{Graph Positional and Structural Encoder}
\usepackage[scaled=.9]{beramono} % use a nicer font size for the \texttt{} command that also has bold version.

\begin{document}

\twocolumn[
\icmltitle{Graph Positional and Structural Encoder}



\icmlsetsymbol{equal}{*}

\begin{icmlauthorlist}
\icmlauthor{Semih Cantürk}{equal,udem,mila}
\icmlauthor{Renming Liu}{equal,msu}
\icmlauthor{Olivier Lapointe-Gagné}{udem,mila}
\icmlauthor{Vincent Létourneau}{udem,mila}
\icmlauthor{Guy Wolf}{udem,mila}
\icmlauthor{Dominique Beaini}{udem,mila,valence}
\icmlauthor{Ladislav Rampášek}{isom}
\end{icmlauthorlist}

\icmlaffiliation{udem}{DIRO, Université de Montréal, Montréal, Canada}
\icmlaffiliation{mila}{Mila -- Quebec AI Institute, Montréal, Canada}
\icmlaffiliation{msu}{Department of Computational Mathematics, Science and Engineering, Michigan State University, East Lansing, United States}
\icmlaffiliation{valence}{Valence Labs, Montréal, Canada}
\icmlaffiliation{isom}{Isomorphic Labs, London, UK}

\icmlcorrespondingauthor{Semih Cantürk}{semih.canturk@mila.quebec}
\icmlcorrespondingauthor{Renming Liu}{liurenmi@msu.edu}

\icmlkeywords{Machine Learning, ICML}

\vskip 0.3in
]



\printAffiliationsAndNotice{\icmlEqualContribution} % otherwise use the standard text.

\begin{abstract}
% This document provides a basic paper template and submission guidelines. Abstracts must be a single paragraph, ideally between 4--6 sentences long. Gross violations will trigger corrections at the camera-ready phase.
Positional and structural encodings (PSE) enable better identifiability of nodes within a graph, rendering them essential tools for empowering modern GNNs, and in particular graph Transformers.
However, designing PSEs that work optimally for all graph prediction tasks is a challenging and unsolved problem.
Here, we present the 
%\underline{g}raph \underline{p}ositional and \underline{s}tructural \underline{e}ncoder 
Graph Positional and Structural Encoder (GPSE), 
the first-ever graph encoder designed to capture rich PSE \textit{representations} for augmenting any GNN.
GPSE learns an efficient common latent representation for multiple PSEs, and is highly transferable: The encoder trained on a particular graph dataset can be used effectively on datasets drawn from markedly different distributions and modalities. We show that across a wide range of benchmarks, GPSE-enhanced models can significantly outperform those that employ explicitly computed PSEs, and at least match their performance in others. Our results pave the way for the development of foundational pre-trained graph encoders for extracting positional and structural information, and highlight their potential as a more powerful and efficient alternative to explicitly computed PSEs and existing self-supervised pre-training approaches.
% We integrate GPSE into the PyG library, and make our framework and pre-trained models available\footnote{\url{https://github.com/G-Taxonomy-Workgroup/GPSE}} for easy downstream usage and the training of new GPSE models.
Our framework and pre-trained models are publicly available\footnote{\url{https://github.com/G-Taxonomy-Workgroup/GPSE}}. For convenience, GPSE has also been integrated into the PyG library to facilitate downstream applications.
\end{abstract}

\section{Introduction}

Graph neural networks (GNN)~\cite{scarselli_graph_2009} are the dominant paradigm in graph representation learning~\citep{hamilton2017representation,bronstein2021geometric}, spanning diverse applications across many domains in biomedicine~\citep{yi2022graph}, molecular chemistry~\citep{xia2022pre}, and more~\citep{benchmarking_gnns,hu2020open,hu2021ogb,liu2022taxonomy}. For most of its relatively short history, GNN algorithms were developed within the message-passing neural network (MPNN) framework~\citep{gilmer2017neural}, where vertices exchange internal states within their neighborhoods defined by the typically sparse graph structure.
This standard MPNN framework has several fundamental limits,
% Despite being computationally efficient, the sparsity leveraged by MPNNs have raised many fundamental limits,
such as the 1-WL bounded expressiveness~\citep{xu2018powerful, wl_go_neural}, under-reaching~\citep{barcelo2020logical}, and over-squashing~\citep{alon2021on,topping2022understanding}. Leveraging the success of the Transformer model in natural language processing~\citep{vaswani2017attention}, graph Transformer (GT) models were developed as a new paradigm for GNNs to address the above limitations by attending to all node pairs in a graph~\citep{dwivedi2021generalization}.
% Attending to all pairs of nodes in a graph circumvents the aforementioned sparsity-induced limitations of MPNN, but it also discards all inductive biases relating to the graph structure~\citep{battaglia2018relational}, which MPNNs leverage well.
% Thus, reintroducing such inductive bias via positional and structural encodings (PSE) has been one of the most essential steps that led to the early success of GT~\citep{rampavsek2022recipe,graph_bert,graphit,kreuzer2021rethinking,dwivedi2021generalization,graphormer}.
% Thus, reintroducing such inductive bias via positional and structural encodings (PSE) has been one of the most essential steps that led to the early success of GT~\citep{rampavsek2022recipe,dwivedi2021generalization,graphormer}.
This full attention inevitably discards the inductive biases related to the graph structure~\citep{battaglia2018relational}, which MPNNs leverage well to excel.
Consequently, positional and structural encodings (PSE) have played the quintessential role in reintroducing such inductive biases, leading to the remarkable success of GTs~\citep{rampavsek2022recipe,dwivedi2021generalization,sat_chen_2022, graphormer}.%,graphormer}.

While many different types of hand-crafted PSEs have been proposed in the literature and used by various GT models, there is no \textit{one-size-fits-all PSE} that performs optimally for all tasks. For example, random walk encodings are more effective for molecular property prediction tasks~\citep{rampavsek2022recipe,lpse}. Conversely, graph Laplacian eigenvectors are more useful for tasks involving long-range dependencies~\citep{dwivedi2022long,benchmarking_gnns}. Moreover, naively stacking different PSEs together does not yield the expected gains. As a result, researchers have to rely on a combination of heuristics, trial-and-error and domain know-how to select the single best encoding for their respective tasks. Developing a universal encoding that combines the benefits of diverse hand-crafted PSEs is thus fundamental to bringing the most performance out of existing GT models.
% Therefore, developing a systematic approach to learn a universal encoding that integrates various PSEs remains an open challenge.
% While many different types of PSEs have been hand-crafted and used by various GT models, a consensus on the best PSE to use in a general context has yet to be reached. In fact, the optimal choice of PSE often depends on the specific task at hand. For example, random walk encodings are typically effective for small molecular tasks~\citep{rampavsek2022recipe,lpse}, while graph Laplacian eigenvectors might be more suitable for tasks that involve long-range dependencies~\citep{dwivedi2022long,benchmarking_gnns}. Therefore, developing a systematic approach to learn a unified encoding that integrates various PSEs remains an open challenge.

%\method shows that we’re at a point where computing and trial-erroring PSEs for each graph is not necessary anymore: We usher a new paradigm where pretrained foundation models (probs not the right term but) can be used for general-purpose PSE generation as opposed to computing and trial-erroring multiple PSEs. It also alleviates some requirements on finding “more expressive architectures”.

Here, we present \method, an MPNN that is trained to extract graph encodings as latent representations of diverse PSEs. Once trained on a pre-training graph dataset by learning to reconstruct different PSEs using only the graph structures, \method can then extract PSE representations from \textit{any} graph dataset to augment GT models. However, designing an MPNN that extracts PSE representations effectively poses a fundamental challenge: \textit{Ensuring that the MPNN suffices to capture properties necessitated by all target PSEs.} First, encodings derived from random walks require beyond 1-WL expressivity~\citep{li2020distance}, while standard MPNNs are known to be bound by 1-WL~\citep{xu2018powerful, wl_go_neural}. Second, graph Laplacian eigenvectors, especially those associated with low frequencies, need access to a global view of the graph structure, which simple MPNNs fail to capture~\citep{furer2010power}. Furthermore, naive solutions to obtain global structural information, such as stacking more MPNN layers, suffer from well-known issues of over-smoothing and over-squashing~\citep{alon2021on, li2019deepgcns}. Through careful architectural design, we mitigate the aforementioned pitfalls and successfully achieve the goal of extracting rich representations from diverse PSEs.
% Encoding information that is transferable between certain graph datasets and tasks is typically achieved via self-supervised learning (SSL) methods. Pre-training GNNs via SSL has proven effective in learning graph features in data-abundant settings, which can then be transferred to downstream tasks~\citep{Hu*2020Strategies, wang2022evaluating, xie2022self}. However, SSL on graphs is plagued by a critical drawback: SSL that performs well on one downstream task may not help or even lead to negative transfer on another, and the success of graph SSL methods typically hinges on whether the data for the pre-training and downstream tasks are well-aligned~\citep{sun2022does, Hu*2020Strategies, wang2022evaluating}.

% Encoding information that is transferable between certain graph datasets and tasks is typically achieved via self-supervised learning (SSL) methods. Pre-training GNNs via SSL has proven effective in learning graph features in data-abundant settings, which can then be transferred to downstream tasks~\citep{Hu*2020Strategies, wang2022evaluating, xie2022self}. However, SSL on graphs is plagued by a critical drawback: SSL that performs well on one downstream task may not help or even lead to negative transfer on another, and the success of graph SSL methods typically hinges on whether the data for the pre-training and downstream tasks are well-aligned~\citep{sun2022does, Hu*2020Strategies, wang2022evaluating}.

\method represents a leap forward in building foundational graph encoders as a one-stop shop for extracting general-purpose PSEs from any graph, alleviating the burden of trial-and-error-based feature engineering associated with conventional PSEs. By demonstrating that \method is computationally efficient and highly performant on a large variety of tasks, we usher in a new paradigm in graph learning without manual PSE engineering and open up exciting opportunities toward more powerful PSE extractors.

We summarize our main contributions as follows.
% In this work, we aim to address the problem of non-generalizable and sensitive PSEs and that of unreliable pre-trained encodings of graph SSL simultaneously by training a \underline{g}raph \underline{p}ositional and \underline{s}tructural \underline{e}ncoder (\method). We summarize our main contributions as follows:

\begin{enumerate}[topsep=-1pt,itemsep=0pt,partopsep=0pt,parsep=3pt,leftmargin=2em,itemindent=0em]
    \item We propose \method, the first attempt at training a foundation graph encoder that extracts rich \textit{positional and structural representations} solely from graph structures, which can be applied to any MPNN or GT model as a replacement for explicitly constructed PSEs.
    \item We show that \method provides significant performance improvements over traditional hand-crafted PSEs across a variety of benchmarks.
    \item Through extensive experiments, we demonstrate that \method is highly \textit{transferable} across graphs of different sizes, connectivity patterns, and modalities.
\end{enumerate}

% \vspace{-0.1in}
\subsection{Related work}

Several approaches have been proposed to overcome the aforementioned limitations of standard MPNNs: \citet{wl_go_neural} propose higher-order MPNNs, \citet{gutteridge2023drew} and \citet{barbero2024localityaware} optimize information flow on graphs through graph rewiring, while \citet{ding2024recurrent} draw inspiration from deep state-space models and RNNs.
%A vast amount of recent work, meanwhile, has focused on variations of graph Transformers enhanced with positional and structural encodings.

% \vspace{-0.1in}
% \textbf{Positional and structural encodings (PSE) }
\textit{Positional encodings} were originally implemented as a series of sinusoidal functions in the Transformer model to capture the ordinal position of words in a sentence~\citep{vaswani2017attention}. However, capturing the positions of nodes in a graph is harder, as nodes of a graph lack such a canonical ordering.
Many recent works on graph Transformers (GT) thus use the graph Laplacian eigenvectors as the positional encodings~\citep{rampavsek2022recipe,kreuzer2021rethinking},
which are direct analogues to the sinusoids in Euclidean space~\citep{spielman2012spectral}. Other methods for encoding positional information include electrostatic potential encodings~\citep{kreuzer2021rethinking}, shortest-path distances~\citep{graphormer}, and tree-based encodings~\citep{shiv2019novel}. \textit{Structural encodings}, on the other hand, have been developed to encode rich local and global connectivity patterns on graph-structured data.
The random walk encoding, for example, has shown great success when used with GTs, particularly on small molecular graph benchmarks~\citep{rampavsek2022recipe,dwivedi2021generalization,lpse}.
Other notable structural encodings include the heat kernel~\citep{kreuzer2021rethinking,graphit}, subgraphs~\citep{bouritsas2022improving,zhao2021stars, sat_chen_2022}, and node degree centralities~\citep{graphormer}.

PSEs are also useful for MPNNs, besides GTs. They can be directly used as additional node features for an MPNN% that are combined with the original graph features %before feeding into the models
~\citep{benchmarking_gnns,lim2022sign,wang2022equivariant}. Other works designed approaches to process the PSEs separately from the original node features. For example, LSPE~\citep{lpse} processes the PSEs with a separate channel and applies an auxiliary loss. SignNet and BasisNet explicitly design components in the prediction model to handle the graph Laplacian eigenvectors, aiming to resolve the sign- and basis-ambiguity issues~\cite{lim2022sign}. Despite this great amount of work in developing methods to better utilize hand-crafted PSEs, it is still unclear how to systematically encode information from multiple types of PSEs to effectively augment GNNs for diverse applications. \method thus represents the first of its kind in an attempt to tackle this fundamental problem.

\begin{figure*}[!htp]
    \centering
    \includegraphics[width=\linewidth]{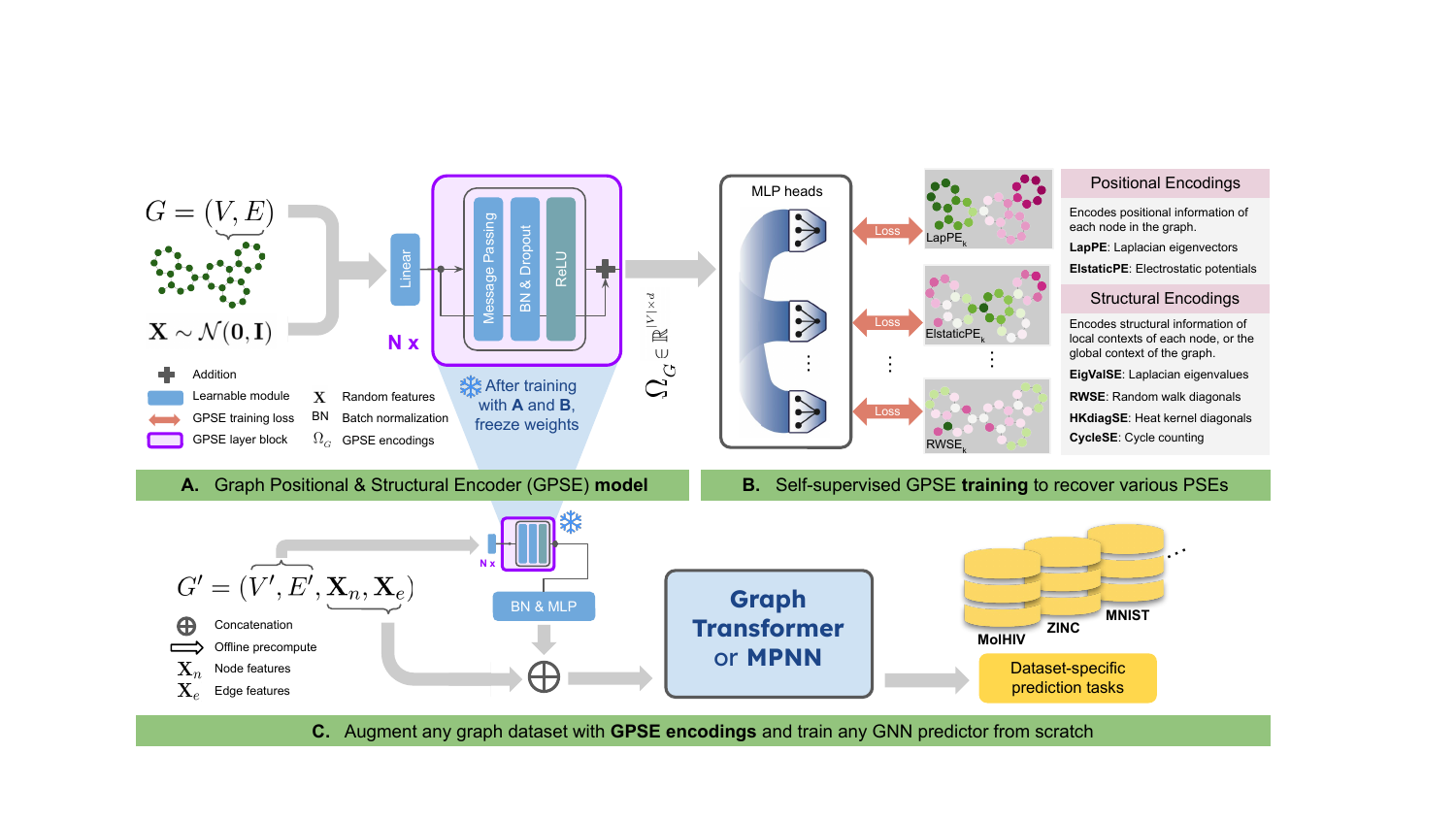}\vspace{-8pt}
    \caption{Overview of Graph Positional and Structural Encoder (\method) training and application.}
    \label{fig:abs}
    \vspace{-5pt}
\end{figure*}

%%%%%%%%%%%%%%%%%%%%%%%%%%%%
\section{Methods}

Our core idea is to train an MPNN as a graph encoder to extract rich positional and structural representations of any query graph based \textit{solely} on its graph structure (Figure~\ref{fig:abs}A). To achieve this, we design a collection of PSEs encompassing a broad range of encodings and use them as self-supervision to train the encoder via reconstruction (Figure~\ref{fig:abs}B). Once the encoder is trained, it can then be used in inference mode to extract PSE representations for augmenting any downstream dataset (Figure~\ref{fig:abs}C).

For downstream tasks, we primarily use the powerful graph Transformer model GPS~\citep{rampavsek2022recipe} that leverages the advantages of both the inductive bias of the local message passing~\citep{battaglia2018relational} and the expressiveness of the global attention~\citep{vaswani2017attention}. As it has previously attained SOTA results on a variety of benchmarks using hand-crafted PSEs, GPS is a natural baseline model to demonstrate the effectiveness of \method. 
We also validate \method for other graph Transformers \& MPNNs in our experiments, and thus show that utility of \method is not bound to any particular architecture.

\subsection{Self-supervision via positional and structural encodings (PSE)} \label{sec:target_pse}

We design a diverse collection of six PSEs for \method to learn against, including the Laplacian eigenvectors (4) and eigenvalues (4), the electrostatic positional encodings (7), the random walk structural encodings (20), the heat kernel structural encodings (20), and the cycle counting graph encodings (7). In short, positional encodings inform the relative position of each node in the graph, while structural encodings describe the local connectivity patterns around a node (Figure~\ref{fig:abs}B). 
See Appendix~\ref{sec:posenc} for precise mathematical definitions of all PSEs used in \method training.

\subsection{\method architecture}\label{sec:arc}

At a high level, our \method model is an MPNN consisting of stacked graph convolution blocks with residual gating, and skip-connections in-between. The mathematical formulation of the architecture can be found in Appendix~\ref{appendix:gpse_comp}. We illustrate below that this careful design enables \method to excel at learning powerful PSE representations. %These components are chosen to minimize two formulations of information bottlenecks in the message-passing process, \emph{over-smoothing} and \emph{over-squashing}. 
%In this section, e hereby discuss these bottlenecks in the form of and the architectural components used in \method to overcome them.
%\textcolor{blue}{Over-smoothing and over-squashing are two key challenges to overcome when learning with deep MPNNs. Thus, our \method model also has to overcome these challenges to be able to learn good joint representations for PSEs.} 
Particularly, we present different architectural choices and \emph{how} they remedy the two key challenges of \emph{over-smoothing} and \emph{over-squashing} (see Appendix~\ref{appendix:theory_details} for technical details).% mathematical formulations of the two phenomena and theoretical justifications behind our design in Appendix~\ref{appendix:theory_details}. %Please refer to Appendix~\ref{appendix:gpse_comp} for a detailed step-by-step analysis of the \method layer, and Appendix~\ref{appendix:theory_details} for technical definitions of over-smoothing and over-squashing; We also discuss their relevance to learning a powerful \method model in further detail in Appendix~\ref{appendix:theory_details_oo_gpse}.

% \todo{See if we can shorten, move tangentially related stuff to appendix or simply remove.}
Evidence suggests that over-smoothing and over-squashing in graph networks relate to the graph's curvature~\citep{topping2022understanding,nguyenRevisitingOversmoothingOversquashing2023,giraldoUnderstandingRelationshipOversmoothing2022}. Formulations of graph curvature~\citep{formanBochnerMethodCell2003, OLLIVIER2009810,sreejithFormanCurvatureComplex2016,topping2022understanding} aim to encode how a node's neighborhood looks like a clique (positive curvature), a grid (zero curvature), or a tree (negative curvature). A clique's high connectivity leads to rapid smoothing, while a tree's exponentially increasing size of $k$-hop neighborhood causes over-squashing. These phenomena are in competition, and negating both is impossible in an MPNN using graph rewiring (architectures using modified adjacency for node aggregation). However, there seems to be a \textit{sweet spot} where the two effects are not minimized by themselves, but their sum is minimized. This minimum is sought in~\citet{giraldoUnderstandingRelationshipOversmoothing2022} and \citet{nguyenRevisitingOversmoothingOversquashing2023}. Some of our following choices of architecture are justified by this search of a sweet spot in the smoothing-squashing trade-off.

\textbf{Deep GNN }
 As several of the target PSEs, such as the Laplacian eigenvectors, require having a global view of the query graph, it is crucial for the encoder to capture long-range dependencies accurately. To accomplish this, we need to use an unconventionally deep MPNN with 20 layers. However, if a graph network suffers from over-smoothing, having this many layers will result in approximately uniform node features ~\citep{oono2019graph,li2019deepgcns}.
 
\textbf{Residual connections \& gating mechanism }
 A first attempt at reducing the smoothing is to exploit the proven ability of residual connections in reducing over-smoothing~\citep{li2019deepgcns}. Using a gating mechanism in aggregation helps reduce the over-smoothing even further. Indeed, gating allows the network to reduce the weight of some edges and in the limit effectively re-wire the graph by completely or partially ignoring some edges. We argue that it is possible for gating to act as a graph sparsification device, which decreases the graph curvature and have been shown to alleviate over-smoothing~\citep{giraldoUnderstandingRelationshipOversmoothing2022, Rong2020DropEdge:}.

\textbf{Virtual node }
In addition, we use a virtual node (VN)~\citep{gilmer2017neural} to enable global message passing; as the virtual node has access to the states of all nodes, it allows for (a) better representation of graph-level information and (b) faster propagation of information between nodes that are further apart, and thus faster convergence of states. In technical terms, adding the virtual node drastically increases the connectivity of the graph and in turn its curvature (Appendix~\ref{appendix:theory_details}, Prop.~\ref{prop:VNincreasesCurvature}), and consequently decreases over-squashing. Alternatively, one can see that the Cheeger constant (another measure of \textit{bottleneckness}~\citep{topping2022understanding}) of the graph increases after adding the virtual node.
% \\ We argue that this device, along with the gating mechanism, likely allows the network to perform a simple search for the aforementioned sweet spot in the graph curvature, thus minimizing the sum of the effects of the over-smoothing and squashing. We later evaluate the elevated performance due to these architecture choices (\S\ref{sec:abla}).

% \textbf{Enabling powerful MPNN via random node features}
\textbf{Random node features }
One critical question is whether an MPNN is expressive enough to learn all the target PSEs. In particular, some PSEs, such as the Laplacian eigenvalues, may require distinguishability beyond 1-WL~\citep{furer2010power}. 
%For example, the multi-sets of eigenvalues of two triangles versus a hexagon are different; this graph pair is commonly known to be 1-WL indistinguishable. 
Despite the known 1-WL expressiveness limitation of a standard MPNN when using constant node features~\citep{xu2018powerful, wl_go_neural}, random node features can help MPNNs surpass 1-WL expressiveness \citep{sato2021random,abboud2020surprising,kanatsoulis2022graph}. Thus, we base our encoder architecture on an MPNN coupled with random input node features, as shown in Figure~\ref{fig:abs}A.
% We later empirically validate the elevated expressiveness due to random features in the graph isomorphism benchmarking experiments (\S\ref{sec:expr}).

We empirically validate that together, the above architectural design choices lead to an effective graph encoder that finds the balance between smoothing and squashing (\S\ref{sec:abla}), and even has an elevated expressiveness due to random features (\S\ref{sec:expr}). A detailed ablation study highlighting the importance of our architectural choices is available in Table~\ref{tab:abl_pse_perf}.

\subsection{Training \method}

Given a query graph structure $G = (V, E)$, we first generate a $k$-dimensional feature from a standard normal distribution for each node, $\mathbf{X} \sim \mathcal{N}(\mathbf{0}, \mathbf{I})$, which is then passed through the \method model to extract the final representations (Figure~\ref{fig:abs}A).
% a linear projection layer to match the $d$-dimensional hidden dimension. Then, the projected features and the graph structure are processed by $N$ message-passing layers with residual connections~\citep{li2019deepgcns}, resulting in the final \method representations Figure~\ref{fig:abs}. 
The representations are then decoded into the target PSEs using multiple independent MLP heads, one per PSE (Figure~\ref{fig:abs}B). We compute the reconstruction loss based on the sum of $\ell_1$ and cosine similarity losses (Appendix~\ref{appendix:loss}), and optimize \method by minimizing this reconstruction loss. Details about hyperparameters can be found in Table~\ref{tab:training_params}.

% This learning approach falls into the category of predictive learning. We note that contrastive approaches are infeasible for learning PSE representations as it is undesirable to obtain a representation that is insensitive to structural perturbations.
% \todo{briefly mention why \method is not feasible for contrastive pre-training}

\textbf{Loss function } The aforementioned combination of $\ell_1$ and cosine similarity losses ensures that the model captures both (1) the \emph{direction} of a particular PSE as a signal on the graph (via cosine similarity loss), and (2) the \emph{magnitude} of the PSE (via $\ell_1$ loss). 
From a graph signal processing perspective, both types of information are crucial to describe the characteristics of PSEs in the form of a graph signal. 
% A more detailed formulation of the loss can be found in Appendix \ref{appendix:loss}.

\textbf{Training dataset }
PCQM4Mv2~\citep{hu2021ogb} is a typical choice of pre-training dataset for molecular tasks. However, since \method only extracts features from graph structures (e.g., methane, CH4, would be treated as the same graph as silane, SiH4), the amount of training samples reduces to 273,920 after extracting unique graphs. Instead, we train \method with MolPCBA~\citep{hu2020open} with 323,555 unique molecular graphs and an average number of 25 nodes. We randomly select 5\% validation and 5\% testing data fixed across runs, and use the remaining data for training \method. An ablation study on training datasets considered is available in Table~\ref{tab:abl_pt_ds}.

\input{tables/pse_perf}

\section{Experiments}

% \lr{Dataset names are inconsistently font-typed with 'texttt', don't use it, only keep that font for \method or make another clear and consistent choice.}

\input{tables/mol_bench}

% \subsection{\method successfully predicts a wide range of target PSEs}
\paragraph{\method successfully predicts a wide range of target PSEs}

The self-supervised goal of \method is to learn graph representations from which it is possible to recover predefined positional and structural encodings. For each PSE type, we quantify prediction performance in terms of the coefficient of determination (\Rsq) scores, as presented in Table~\ref{tab:pse_perf}. When trained on a 5\% (16,177) subset of MolPCBA molecular graphs, \method achieves 0.9790 average test \Rsq score across the 6 PSEs. 
Further, we show that test performance improves asymptotically as the number of training samples increases (\S\ref{sec:asym}), achieving 0.9979 average test \Rsq when trained on 90\% (291,199) of unique MolPCBA graphs. These results demonstrate the ability of \method to extract rich positional and structural information from a query graph, as well as its ability to learn from increasing amount of data.

% \lr{May be list in parenthesis the actual number of graphs corresponding to 5\% and 90\% of MolPCBA.}
% \todo{Add cross dataset PSE evaluations to appendix if available}

% - Trained using 5\% data on ZINC, with 5\% holdout validation and testing.
% - \method successfully captures positional and structural information with overall all R2 or over 0.98.
% - The good performance comes from multiple engineering and design choices of the architecture: multi-MLP head, width and depth, GatedGCN that goes deep, VN to capture global information.

\subsection{Enhancing performance on molecular graph data}

In these experiments, we demonstrate that \method provides more performance improvements over traditional PSEs for a wide range of GNN models. Additionally, we show competitive performance achieved by \method against the complementary self-supervised learning (SSL) pre-training approaches.

% \paragraph{Molecular benchmarking graph datasets}
\textbf{\method-augmented GPS is highly competitive on molecular graph benchmarks }
% \lr{This needs to be restructured a little (after the Transformer results are in), how to name\&pitch the paragraph that is about Table~\ref{tab:molbench} and the one about Table~\ref{tab:abl_zinc_pse}}
We compare the performance of the GPS model augmented with our \method encodings versus the same model using (a) no PSE, (b) random features as PSE, (c) LapPE and RWSE, and (d) concatenation of PSEs from \S\ref{sec:target_pse} on four common molecular property prediction benchmarks~\citep{benchmarking_gnns,hu2020open,hu2021ogb}. For ZINC~\citep{gomez2018automatic} and PCQM4Mv2~\citep{hu2020open}, we use their subset versions following~\citet{benchmarking_gnns} and~\citet{rampavsek2022recipe}, respectively.
% \texttt{ZINC-subset} and \texttt{PCQM4Mv2-subset}
% We also experiment with appending the final latent representation of our PSEs (before the per-PSE MLPs) instead of the predicted ones themselves. The \method-generated PSEs not only match but surpass both GPS+LapPE and GPS+RWSE; we also find that appending the final latent representation directly is better than using the predicted PSEs, it is likely that this representation contains additional information with respect to the combination of the individual PSEs.

We first highlight that \method-augmented GPS achieves a remarkable 0.0648 MAE on ZINC (Table~\ref{tab:molbench}), not only significantly outperforming other PSEs but even challenging SOTA results. Similarly, GPS+\method also achieves the best result on PCQM4Mv2 amongst models that (a) are not ensemble methods and (b) do not have access to 3D information. These two strategies are highly engineering- and domain-oriented, thus do not reflect their utilities on general graph learning tasks, which we aim to demonstrate instead.
% \todo{the starting of the last sentence reads a bit weird. also need to give more context about the two criterions for readers - the leaderboard stuff}

Moreover, we note that \method always performs better than, or at least on par with, standard PSEs, while concatenating multiple PSEs (AllPSE/LapPE+RWSE) always leads to worse performance than using individual PSEs. On ZINC \& PCQM4Mv2, \method improves results comfortably beyond standard deviation, while in the worst case (e.g., MolHIV) it recovers the best PSE result, with differences well within a standard deviation. We discuss why \method works better on some datasets than others in Appendix~\ref{appendix:discussion}.

% - Improved performance over both LapPE and RWSE augmented GPS on the two molecular benchmarking datasets.
% - Note that the  PCQM4Mv2 subset performances differ from that reported in~\citep{rampavsek2022recipe} possibly due to randomness in the subsampling (\texttt{numpy} random generator is not guaranteed to be persistent across \texttt{numpy} versions). We will publish the exact split indices we used in this work.

\textbf{\method as a universal PSE augmentation }
The utility of \method encodings is not specific to GPS. In Table~\ref{tab:molbench}, we show that \method also significantly enhances two variants of SAT~\citep{sat_chen_2022} on ZINC. Additionally, we show that augmenting different MPNN methods and the graph Transformer with \method universally results in remarkable improvements on ZINC: 56.24\% reduction in test MAE on average compared to baselines that do not make use of any PSE, outperforming all other PSEs and their combinations (Table~\ref{tab:abl_zinc_pse}). 
%This improvement is the most significant compared with using a single or concatenated PSEs. 
Notably, concatenating LapPE and RWSE does not yield any benefit beyond using either one of them, demonstrating the intricacy of effectively leveraging information from multiple PSEs. We additionally perform a similar set of experiments on PCQM4Mv2 (Table~\ref{tab:abl_pcqm4m_pse}), and obtain the same improvements over explicitly computed PSEs, validating the success of \method further.

\input{tables/abl_zinc_pse}

\textbf{Feature augmentation using \method vs.~SSL pre-training }
Our \method feature augmentation is related to self-supervised learning (SSL) pre-training approaches~\citep{Hu*2020Strategies,you2020graph,xie2022self,xu2021self} in that both transfer knowledge from a large pre-training dataset to another for downstream evaluation. However, our approach is a substantial departure from previous SSL approaches in two distinct aspects:

\begin{enumerate}[topsep=0pt,itemsep=0.5ex,partopsep=0pt,parsep=0.25ex,leftmargin=2em,itemindent=0em]
    \item The trained \method model only serves as a feature extractor that can be coupled with \textit{any} type of downstream prediction model, which will be trained from scratch.
    \item \method extracts representations \textit{solely} from the graph structure and does not make use of the domain-specific features such as atom and bond types~\citep{Hu*2020Strategies}, allowing \method to be utilized on any graph dataset.
\end{enumerate}

% \lr{seems confusing, two questions: 1) didn't we establish general usefulness in the prior sections already? isn't the goal explicitly to compare to SSL methods? 2) the way \citep{sun2022does} and \citep{Hu*2020Strategies} are referred to in the first sentence is confusing.}
% To study the usefulness of \method as a general feature augmentation, we follow a comprehensive ablation study~\citep{sun2022does} on the effectiveness of SSL pre-training using a selection of five MoleculeNet~\citep{wu2018moleculenet} datasets from~\citep{Hu*2020Strategies}.
To compare the performance of SSL pre-training and \method feature augmentation, we
% follow~\citet{sun2022does} and use a selection of eight
use the MoleculeNet~\citep{wu2018moleculenet,Hu*2020Strategies} datasets.
For the downstream model, we use the identical GINE architecture \citep{Hu*2020Strategies} from~\citet{sun2022does}.
%, which is an edge attribute aware variant of GIN, with five layers and 300-wide hidden dimensions. 
Finally, the extracted representations from \method are concatenated with the atom embeddings and are then fed into the GINE model.

\input{tables/molnet}

We note that \method-augmented GINE achieves the best performance on three out of the eight MoleculeNet datasets against previously reported performances (Table~\ref{tab:molnet}). Moreover, \method augmentation improves performance over the baseline across \textit{all} eight datasets, unlike some of the previously reported results that show negative transfer. Together, these results corroborate with the findings from~\citet{sun2022does} that rich features can make up for the benefits of SSL pre-training. In our case, the \method encodings act as the rich features that contain positional and structural information from the graphs.

We also highlight that the Table~\ref{tab:molnet} results are achieved in a setup where \method is at a comparative disadvantage: As a general-purpose feature extractor trained on a separate dataset, \method cannot leverage atom and bond features of the downstream graphs unlike typical molecular graph SSL methods. When GraphLoG~\citep{xu2021self} is similarly used as a feature extractor only, for a fair comparison, it is well outperformed by \method and even suffers from negative transfer, highlighting the power of \method as a feature extractor. With this in mind, \method can potentially be combined with other SSL methods to enhance them in future work.%since we have shown its ability to improve the expressivity of GINE. 

\subsection{Transferability across diverse graph benchmarks}

\method can be used on arbitrary types of graphs as it is trained using the graph structures alone, in contrast to the SSL pre-training methods.
Here, we show that \method is transferable to general graph datasets beyond molecular data, even under extreme out-of-distribution (OOD) cases.

%\input{tables/graph_bench}

% \paragraph{Augmenting long range and global feature extraction}
\textbf{Transferability to molecular graph sizes }
We use Peptides-struct and Peptides-func from the Long Range Graph Benchmark~\citep{dwivedi2022long} to test whether \method can still work when the downstream (macro-)molecular graphs are substantially larger than those used for training.
% Specifically, MolPCBA graphs contain 25 nodes, while Peptides graphs contain 150 nodes, on average.
Despite this difference in graph sizes, \method outperforms explicitly computed PSEs when used with GPS as well as a GCN architecture optimized for long-range benchmarks by \citet{tönshoff2023did} (Table~\ref{tab:graphbench}). More strikingly, GPS+\method challenges SOTA results for Peptides-struct, surpassing Graph MLP-Mixer~\citep{he2022generalization}. The improved performance emphasizes the ability of \method to better extract global information from query graphs by providing a more informative initial encoding for the global attention mechanism in GPS.

\textbf{Transferability to graph connectivity patterns }
We further test if \method generalizes to graph connectivity patterns distinct from its training dataset. Particularly, we use the superpixel graph benchmarking datasets CIFAR10 and MNIST from \citet{benchmarking_gnns}, which are $k$-nearest neighbor graphs with $k\!=\!8$, and employ connectivity patterns that significantly differ from those found in molecular graph datasets. Impressively, GPS+\method again achieves comparable results against GPS + computed PSEs (Table~\ref{tab:graphbench}).
%  and employ connectivity patterns significantly different from those of molecular graphs

\input{tables/graph_bench}

% \paragraph{Out of distribution graph benchmarks}

% The peptides (Peptides-struct and Peptides-func)~\citep{dwivedi2022long} and the superpixel (CIFAR and MNIST) datasets~\citep{benchmarking_gnns} consist of graphs with significantly different structural characteristics from the training dataset MolPCBA for \method. In particular, while MolPCBA contains molecule graphs with 25 nodes on average, the graphs in the peptides and superpixel datasets are much larger (see Appendix \todo{???} for detailed dataset stats), with distinct connectivity patterns, for example, the nodes in a superpixel graph have a constant number of neighbors.

% Despite this difference in the graph characteristics, GPS+\method achieves performance no worse than the original GPS that uses explicitly computed PSEs. More strikingly, GPS+\method resulted in the new SOTA performance for Peptides-struct, surpassing Graph MLP-Mixer~\citep{he2022generalization}. The improved performance of GPS due to \method emphasizes its ability to help better extracts global information from the query graphs by providing a more informative initial encoding for the global attention mechanism embedded in the GPS model.

\textbf{Transferability to extreme OOD node-classification benchmarks }
Taking the evaluation of \method one step further, we test its ability to provide useful information to transductive node classification tasks, where the graphs contain hundreds of thousands of nodes, which are completely out of distribution from the \method training dataset. We use both MPNN (GCN, GraphSAGE~\citep{hamilton2017inductive}, GATv2~\citep{velivckovic2017graph,brody2021attentive}) and graph Transformer baselines. Remarkably, on 8/10 model-dataset combinations, \method attains the best or equal-best result, as well as the best overall results, while LapPE performs below the no-encoding baseline for MPNNs (Table~\ref{tab:nodebench_full}).

Meanwhile, the indifference in performance on the Proteins dataset for MPNNs is not unexpected, as the connectivity structures of the protein interaction network do not contribute to the proteins' functions as meaningfully. Instead, the identity of the proteins' interacting partners is of importance, commonly referred to as \textit{homophily} in the graph representation learning community~\citep{zhu2020beyond} or more generally known as the \textit{Guilt-by-Association} principle in network biology~\citep{cowen2017network}. This result provides valuable insights into the usefulness of \method as an augmentation: It is more beneficial when the underlying graph structure is informative for the downstream tasks.

\input{tables/node_bench_new}
\input{tables/synthetic}
\subsection{Expressiveness of \method encodings}\label{sec:expr}
% \vspace{-3pt}
Given that \method can recover different PSEs so well (Table~\ref{tab:pse_perf}), it is natural to wonder whether it boosts standard MPNN expressiveness.
We first confirm that \method encodings surpass 1-WL distinguishability by observing a clear visual separation of \method encodings on 1-WL indistinguishable graph pairs (Figure~\ref{fig:vis_1wl_graphs}). More information regarding graph isomorphism, the WL test and their connections to GNN expressivity is discussed in Appendix~\ref{appendix:wl}.

To rigorously and systematically illustrate the expressivity of \method encodings further, we employ two synthetic benchmarks~\citep{benchmarking_gnns,abboud2020surprising}
that require beyond 1-WL power, and use an MPNN model with 1-WL expressivity in GIN. Indeed, we find that \method provides extra power to the base MPNN to correctly distinguish graph isomorphism classes (Table~\ref{tab:synth}).
This expressivity boost by \method is particularly impressive, considering that (1) \method is pre-trained on MolPCBA, whose graph structures are not designed to be 1-WL indistinguishable like these synthetic graphs, and (2) naively adding random features to the input does not provide the same improvement.

% We stress that this result does not contradict the well-known 1-WL expressivity limitation of MPNNs. Rather, it is a good example of how random features input is a simple yet powerful approach to breaking symmetries and subsequently allowing MPNNs to extract richer features~\citep{abboud2020surprising,sato2021random}.

We further point out that, in fact, augmenting the base GIN model using common PSEs like LapPE and RWSE readily archives nearly perfect graph isomorphism classification, corroborating with previous theoretical results on distance-encodings~\citep{li2020distance} and spectral invariants~\citep{furer2010power}.
% with the fact that spectral invariants (e.g., multisets of the graph Laplacian eigenvalues) are not contained within the 1-WL distinguishing power~\citep{furer2010power}.
This finding partially explains why \method provides additional power and also why previous methods using LapPE achieve perfect classification on these tasks~\citep{he2022generalization}. Finally, we note the performance difference between LapPE/RWSE and \method is not unexpected, as random input features only act as a patch to the MPNN expressivity limitation, rather than fully resolving it. Thus, developing more powerful and practically scalable \method models that losslessly capture the latent semantics of various PSEs is a vital avenue to explore in the future.

\vspace{-2pt}
\subsection{Efficiency \& scaling of \method}\label{sec:scaling}
\vspace{-2pt}
We demonstrate the efficiency and scalability advantages of \method over hand-crafted PSEs through two sets of scaling experiments. In the first, we compare the computation time for different PSEs against \method as we vary the number of graphs, represented as a percentage of the MolPCBA dataset. In the latter set of experiments, we instead generate four datasets of 1000 synthetic graphs, but scale up the individual graph sizes in each dataset. Our results are presented in Appendix~\ref{appendix:scaling}, where we show that \method is not only considerably faster to compute than explicitly computing PSEs, but also scales much better than explicit PSE computation as both the number of graphs and graph sizes increase. 

The benefit of \method further compounds and leads to orders-of-magnitude faster computation times when we compute and combine all PSEs required for a fair comparison (AllPSE). One primary computational advantage of \method is that its complexity remains unchanged at inference time, regardless of the number of PSE types used to train the model. In our study, we restricted ourselves to six different PSEs, but future work could easily include more complex and specialized PSEs yet claim the same efficiency properties for PSE extraction. However, for AllPSE, the computational costs of stacking an increasing number of encodings would quickly become untenable.
%but it is straightforward for future work to include even more complex and specialized PSEs and claim the same efficiency properties for PSE extraction, while for AllPSE the computational costs of stacking an increasing number of encodings would quickly become untenable.

\begin{figure}
    \centering
    \includegraphics[width=\linewidth]{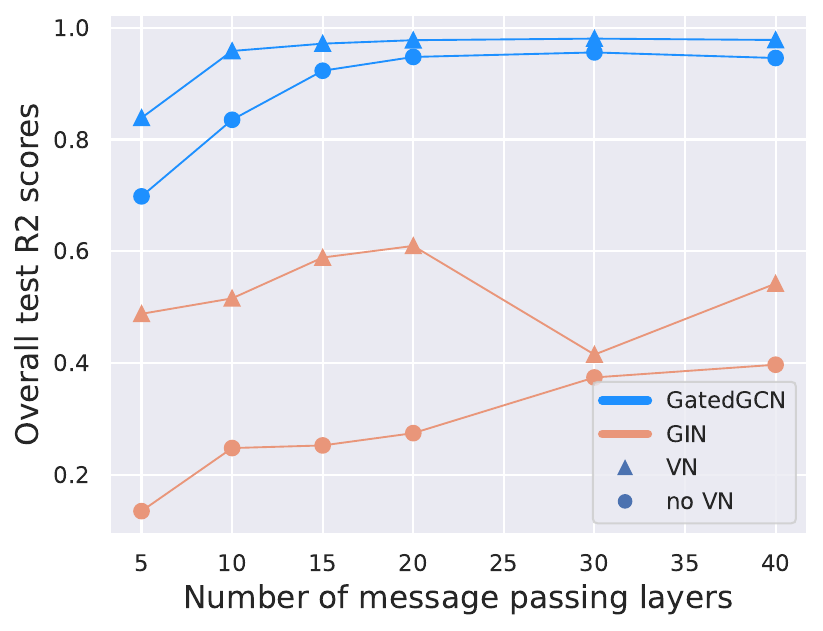}\vspace{-8pt}
    \caption{Virtual node (VN), convolution type, and layers ablation using 5\% MolPCBA for training \method. The y-axis denotes the GPSE average test $R^2$ score over all six PSEs, as per Table~\ref{tab:pse_perf}.}
    \label{fig:abl_conv_layers_vn}
\end{figure}

\subsection{Ablation studies}\label{sec:abla}

% \paragraph{Does \method make good use of the depth and the global message passing from the VN?}
\paragraph{\method makes good use of depth and global message passing from the VN }
Despite the commonly known issue with MPNN over-smoothing as the number of message passing layers increases, we observe that \method does not oversmooth thanks to the gating mechanism and the residual connections in GatedGCN~\citep{bresson2017residual}, benefiting from both the global message passing by VN and the model depth as indicated in Figure~\ref{fig:abl_conv_layers_vn} and \S\ref{sec:arc}.

% \paragraph{Are all target PSE used for training \method helpful for downstream tasks?}
\textbf{Downstream tasks benefit from the wide variety of PSEs used in \method pre-training }
Since \method is trained to capture latent semantics for recovering a wide range of PSEs, it mitigates the reliance on manually selecting task-specific PSEs, a major shortcoming of graph Transformers such as GPS and Graphormer~\citep{graphormer}.
%, which could be task specific. 
For instance, RWSE typically performs well for molecular tasks, while LapPE could be more useful for long-range dependency tasks~\citep{rampavsek2022recipe,dwivedi2022long}. In Table~\ref{tab:abl_pt_task}, we investigate whether a particular type of PSE contributes more or less to \method by testing the downstream performance of PCQM4Mv2 and MolHIV using different variants of \method that excludes one type of PSE during training. We observe that excluding any type of PSE generally reduces its performance in the downstream tasks, indicating the usefulness of different PSEs' semantics to the downstream tasks at various levels.

\textbf{Asymptotic behavior with respect to the \method training sample sizes }\label{sec:asym}
We perform a scaling law experiment with respect to the training sample sizes, from 5\% to 80\% of the MolPCBA. As shown in Figure~\ref{fig:scaling_law}, the test loss (Appendix~\ref{sec:impl}) reduces as the size of the training set increases. This asymptotic behavior suggests that \method can further benefit from the increasing amount of training data.

% \paragraph{How does changing the \method training dataset affect its downstream performance?}
\textbf{The impact of pre-training dataset choice and fine-tuning on \method performance }
We reevaluate the performance of \method on PCQM4Mv2 and ZINC when trained on several other choices of molecular graph datasets, including GEOM~\citep{axelrod2022geom}, ZINC 250k~\citep{gomez2018automatic}, PCQM4Mv2~\citep{hu2021ogb}, and ChEMBL~\citep{gaulton2012chembl}. On ZINC, \method performance is variable across different training datasets (Table~\ref{tab:abl_pt_ds}). Particularly, training \method on ChEMBL and MolPCBA, two largest datasets here, results in much better performances than using other, smaller datasets. The superior downstream performance achieved using larger pre-training datasets aligns well with our asymptotic results above, where a larger amount of training samples results in a more accurate \method for capturing PSEs and better downstream performance. However, we did not observe the same performance difference in the PCQM4Mv2-subset downstream task, indicating that the training size is not always the most crucial factor for good performance, an observation similar to~\citet{sun2022does}.

Finally, we investigate whether fine-tuning the \method model specifically on the downstream dataset could further improve its downstream performance (Table~\ref{tab:abl_pt_ds}). Similar to the above findings, we see that further fine-tuning \method may help in a task-specific manner, generally providing slight improvements on ZINC but less so on PCQM4Mv2. Together, the fact that using different pre-training datasets (provided they are sufficiently diverse) and further fine-tuning to specific downstream datasets having limited effects on \method performance reemphasize that \method learns \textit{general} and \textit{transferable} knowledge about various PSEs.
% However, we did not observe any noticeable improvements from fine-tuning in under most cases. The indifference in performance

% \todo{Test PSE performance on EXP}

The success of \method as a general-purpose PSE encoding raises important discussion points regarding its effectiveness. Of particular interest is how and why \method outperforms individual and concatenated PSEs, even under significant distribution shifts. In Appendix~\ref{appendix:discussion}, we address several such points to both shed light on several aspects regarding \method's effectiveness, and encourage future work towards a better theoretical foundation of \method-like PSE extractors.

\section{Conclusion}

We have introduced \method, a unifying \underline{g}raph \underline{p}ositional and \underline{s}tructural \underline{e}ncoder for augmenting any graph learning dataset while being applicable to all graph Transformers and message-passing GNNs. 
% We have introduced \method, a unifying \underline{g}raph \underline{e}ncoder designed to learn rich latent representations of \underline{p}ositional and \underline{s}tructural encodings, suitable for augmenting any graph learning dataset while being applicable to all graph Transformers and message-passing GNNs. 
% By learning to predict a diverse collection of PESs from the query graphs in the pre-training phase, \method extracts rich node encodings that are vital for the global attention mechanism to function optimally.
\method extracts rich node encodings by learning to predict a diverse collection of predefined PSEs in the initial self-supervised training phase on a set of unattributed graphs.
We demonstrate the superior performance of \method encodings over explicitly constructed PSEs on a variety of graph learning benchmarks. Furthermore, \method shows great transferability across diverse benchmarking datasets and even challenges the SOTA on the Peptides-struct long-range benchmark, whose graph structures are vastly different from those in the MolPCBA dataset used to train the \method model.
% \todo{Closing sentence highlighting the contribution and the impact of this work.}
% Our study opens up exciting opportunities for learning graph encoders as unified PSE representations. We hope that our work will complement the study of PSEs as a parallel line of work, and motivate future studies on learning graph encoders that further enhance graph analysis.

%Our study opens up exciting opportunities to move beyond hand-crafted PSEs to enhance GNNs, and proposes a powerful, transferable and scalable alternative by learning a graph encoder as a unified PSE extractor in \method. 
Our study proposes a powerful, transferable and scalable alternative to hand-crafted PSEs to enhance GNNs.
We therefore hope our work will motivate a shift from the limitations of PSE-based feature-engineering to developing more powerful encoders and foundation models for PSEs as feature extractors to advance the field of graph learning. 
% something that sells our "niche", as in there are interesting lines of research worth looking into for other researchers on learning PSE representations.
%Our results pave the way for the development of large pre-trained models for extracting graph positional and structural information and highlight their potential as a viable alternative to explicitly computed PSEs, thereby opening up new possibilities for more accurate and effective graph analysis.

\textbf{Limitations and future directions }
Despite the effectiveness of our \method model, it is currently prohibitively large to be trained on graph datasets with over one million graphs efficiently. As \method exhibits data scaling laws, where it asymptotically achieves perfect PSE recovery, it is a promising future direction to make \method more efficient and thus allow it to be trained on billion scale molecular graph datasets~\citep{patel2020savi,irwin2020zinc20}.
% such as SAVI~\citep{patel2020savi} and ZINC20~\citep{irwin2020zinc20}.

\section*{Acknowledgements and Disclosure of Funding}

We would like to thank Michael Galkin for discussions essential to the development of this paper. This work was partially funded by Bourse en intelligence artificielle des Études supérieures et postdoctorales (ESP) 2023-2024
[Semih Cantürk]; Canada CIFAR AI Chair, IVADO (Institut de valorisation des données) grant PRF-2019-3583139727, FRQNT (Fonds de recherche du Québec - Nature et technologies) grant 299376, NSERC Discovery grant 03267, and NSF DMS grant 2327211 [Guy Wolf]. This research was also enabled in part by compute resources provided by Mila (mila.quebec). The content provided here is solely the responsibility of the authors and does not necessarily represent the official views of the funding agencies.

\section*{Impact Statement}
This paper presents work whose goal is to advance the field of Machine Learning. There are many potential societal consequences of our work, none which we feel must be specifically highlighted here.
% We do not foresee immediate societal impact of \method as a general graph encoder.

% - scaling up training samples
% - performance could be related to the characteristics of the pre-training dataset - a common issue with pre-training (see Does GNN Pretrianing ...)
% - More PEs -> pairwise task
% - More expressive encoder (but might hinder scalability) -> need novel solutions for scalable and expressive encoder (maybe some hybrid approach)

% \section*{Acknowledgements}

% \textbf{Do not} include acknowledgements in the initial version of the paper submitted for blind review.

% If a paper is accepted, the final camera-ready version can (and probably should) include acknowledgements. In this case, please place such acknowledgements in an unnumbered section at the end of the paper. Typically, this will include thanks to reviewers who gave useful comments, to colleagues who contributed to the ideas, and to funding agencies and corporate sponsors that provided financial support.


\bibliography{references}
\bibliographystyle{icml2024}

%%%%%%%%%%%%%%%%%%%%%%%%%%%%%%%%%%%%%%%%%%%%%%%%%%%%%%%%%%%%%%%%%%%%%%%%%%%%%%%
%%%%%%%%%%%%%%%%%%%%%%%%%%%%%%%%%%%%%%%%%%%%%%%%%%%%%%%%%%%%%%%%%%%%%%%%%%%%%%%
% APPENDIX
%%%%%%%%%%%%%%%%%%%%%%%%%%%%%%%%%%%%%%%%%%%%%%%%%%%%%%%%%%%%%%%%%%%%%%%%%%%%%%%
%%%%%%%%%%%%%%%%%%%%%%%%%%%%%%%%%%%%%%%%%%%%%%%%%%%%%%%%%%%%%%%%%%%%%%%%%%%%%%%
\newpage
\appendix
\onecolumn

% You can have as much text here as you want. The main body must be at most $8$ pages long. For the final version, one more page can be added. If you want, you can use an appendix like this one.  

% The $\mathtt{\backslash onecolumn}$ command above can be kept in place if you prefer a one-column appendix, or can be removed if you prefer a two-column appendix.  Apart from this possible change, the style (font size, spacing, margins, page numbering, etc.) should be kept the same as the main body.

\counterwithin{figure}{section}
\counterwithin{table}{section}
\counterwithin{equation}{section}
\input{appendix}

%%%%%%%%%%%%%%%%%%%%%%%%%%%%%%%%%%%%%%%%%%%%%%%%%%%%%%%%%%%%%%%%%%%%%%%%%%%%%%%
%%%%%%%%%%%%%%%%%%%%%%%%%%%%%%%%%%%%%%%%%%%%%%%%%%%%%%%%%%%%%%%%%%%%%%%%%%%%%%%

\end{document}

%% file: tables/pse_perf.tex
% \usepackage{booktabs}
\begin{wraptable}{r}{0.3\linewidth}\centering
% \vspace{-1.45cm}
\vspace{-0.7cm}
% \caption{Held-out PSE prediction performance when trained on 5\% MolPCBA (\Rsq $\uparrow$).}
\caption{Held-out PSE prediction performance of \method on 5\% MolPCBA.}
\label{tab:pse_perf}
\vspace{2pt}
\footnotesize
\begin{tabular}{@{}ll@{}}
\toprule
\textbf{PSE} & \textbf{\Rsq $\uparrow$} \\ \midrule
ElstaticPE   & 0.964                         \\
LapPE        & 0.973                         \\
RWSE         & 0.984                         \\
HKdiagSE     & 0.981                         \\
EigValSE     & 0.982                         \\
CycleSE      & 0.977                         \\ \midrule
Overall      & 0.979                         \\ \bottomrule
\end{tabular}
% \vspace{-0.2in}
\end{wraptable}

%% file: tables/mol_bench.tex
\begin{table*}[!htp]\centering
\caption{Performance in four molecular property prediction tasks, averaged over 10 seeds.
% \lr{Make sure each Table and Figure is referenced from the main text at least once. Seems many are not.} \rl{Done}
}\label{tab:molbench}
\footnotesize
% \scriptsize
% \tiny
% \vspace{3pt}
% \hspace*{-0.2cm}
\begin{tabular}{lccccc}\toprule
&\textbf{ZINC} (subset) &\textbf{PCQM4Mv2} (subset) &\textbf{MolHIV} &\textbf{MolPCBA}\\
&\textbf{MAE} $\downarrow$ &\textbf{MAE} $\downarrow$ &\textbf{AUROC} $\uparrow$ &\textbf{AP} $\uparrow$\\
\midrule
GCN~\citep{kipf2017semisupervised} &0.3670 ± 0.0110 &-- &0.7599 ± 0.0119 &0.2424 ± 0.0034 \\
GIN~\citep{xu2018powerful} &0.5260 ± 0.0510 &-- &0.7707 ± 0.0149 &0.2703 ± 0.0023 \\
CIN~\citep{bodnar2021weisfeiler_CIN} &0.0790 ± 0.0060 &-- &\textbf{0.8094 ± 0.0057} &-- \\
CRaWI~\citep{toenshoff2021CRaWl} &0.0850 ± 0.0040 &-- &-- &\textbf{0.2986 ± 0.0025} \\
PR-MPNN\textsubscript{SIM}~\citep{qian2023probabilistically} & 0.0850 ± 0.0020 &-- & 0.7950 ± 0.0090  &-- \\
K-ST SAT+RWSE~\citep{sat_chen_2022} &0.1020 ± 0.0050 &-- &-- &-- \\
K-SG SAT+RWSE~\citep{sat_chen_2022} &0.0940 ± 0.0080 &-- &-- &-- \\
% GCN &0.3670 ± 0.0110 &-- &0.7599 ± 0.0119 &0.2424 ± 0.0034 \\
% GIN &0.5260 ± 0.0510 &-- &0.7707 ± 0.0149 &0.2703 ± 0.0023 \\
% CIN &0.0790 ± 0.0060 &-- &\textbf{0.8094 ± 0.0057} &-- \\
% CRaWI &0.0850 ± 0.0040 &-- &-- &\textbf{0.2986 ± 0.0025} \\
% \midrule
% GINE &0.xxxx ± 0.xxxx &0.xxxx ± 0.xxxx &0.xxxx ± 0.xxxx &0.xxxx ± 0.xxxx \\
% GINE+\method &0.xxxx ± 0.xxxx &0.xxxx ± 0.xxxx &0.xxxx ± 0.xxxx &0.xxxx ± 0.xxxx \\
\midrule
K-ST SAT+\textbf{\method} &0.0830 ± 0.0023 &-- &-- &-- \\
K-SG SAT+\textbf{\method} &0.0873 ± 0.0008 &-- &-- &-- \\
\midrule
GPS+rand &0.8766 ± 0.0107 &0.4768 ± 0.0171 &0.6210 ± 0.0444 &0.0753 ± 0.0045\\
GPS+none &0.1182 ± 0.0049 &0.1329 ± 0.0030 &0.7798 ± 0.0077 &0.2869 ± 0.0012\\
GPS+LapPE &0.1078 ± 0.0084 &0.1267 ± 0.0004 &0.7736 ± 0.0097 &0.2939 ± 0.0016\\
GPS+RWSE &0.0700 ± 0.0040 &0.1230 ± 0.0008 &0.7880 ± 0.0101 &0.2907 ± 0.0028 \\
GPS+LapPE+RWSE &0.0822 ± 0.0040 &0.1273 ± 0.0006 &0.7719 ± 0.0129 &0.2854 ± 0.0029 \\
GPS+AllPSE &0.0734 ± 0.0030 &0.1254 ± 0.0011 &0.7645 ± 0.0236 &0.2826 ± 0.0001 \\
\midrule
GPS+\textbf{\method} &\textbf{0.0648 ± 0.0030} &\textbf{0.1196 ± 0.0004} &0.7815 ± 0.0133 &0.2911 ± 0.0036\\  % v9.5_molpcba_noflag (molpcba)
\bottomrule
\end{tabular}
\end{table*}

%% file: tables/abl_zinc_pse.tex
\begin{table*}[!htp]\centering
\caption{Six PSE augmentations combined with five different GNN models evaluated on ZINC (12k subset) dataset. Performance is evaluated as MAE ($\downarrow$) and averaged over 4 seeds.}\label{tab:abl_zinc_pse}
\vspace{2pt}
\footnotesize
% \scriptsize
% \tiny
% \hspace*{-0.15cm}
\begin{tabular}{lcccccc}\toprule
% &GCN &GatedGCN &GIN &GPS (GINE) &GPS (Transformer) &Avg. reduction \\
&GCN &GatedGCN &GIN &GINE &Transformer &Avg. MAE reduction \\
\midrule
none    &0.288 ± 0.004  &0.236 ± 0.008  &0.285 ± 0.004  &0.118 ± 0.005 &0.686 ± 0.017 &-- \\
\midrule
rand    &1.277 ± 0.340  &1.228 ± 0.012  &1.239 ± 0.011  &0.877 ± 0.011  &1.451 ± 0.002 &N/A \\
LapPE   &0.209 ± 0.008  &0.194 ± 0.006  &0.214 ± 0.004  &0.108 ± 0.008  &0.501 ± 0.145 &\cellcolor[HTML]{fff2cc}21.12\% \\
RWSE    &0.181 ± 0.003  &0.167 ± 0.002  &0.175 ± 0.003  &0.070 ± 0.004  &0.219 ± 0.007 &\cellcolor[HTML]{ffe498}42.75\% \\
LapPE+RWSE	&0.184 ± 0.008 &0.163 ± 0.009 &0.174 ± 0.003 &0.082 ± 0.004 &0.205 ± 0.007 &\cellcolor[HTML]{ffe498}41.32\%\\
AllPSE	&0.150 ± 0.007 &0.143 ± 0.007 &0.153 ± 0.006 &0.073 ± 0.003 &0.190 ± 0.008 &\cellcolor[HTML]{fcdd86}50.85\%\\
\textbf{\method}    &\textbf{0.129 ± 0.003}  &\textbf{0.113 ± 0.003}  &\textbf{0.124 ± 0.002}  &\textbf{0.065 ± 0.003}  &\textbf{0.189 ± 0.016} &\cellcolor[HTML]{faca46}\textbf{56.24\%} \\
\bottomrule
\end{tabular}
\end{table*}

%% file: tables/molnet.tex
\begin{table*}[!htp]\centering
%\vspace{-10pt}
\caption{Performance on MoleculeNet datasets (scaffold split), evaluated in AUROC (\%) $\uparrow$. \textcolor{dark2pink}{Red} indicates worse than baseline performance.}\label{tab:molnet}
\vspace{2pt}
\fontsize{7.9pt}{7.9pt}\selectfont
\renewcommand{\arraystretch}{1.1}
\setlength\tabcolsep{4pt} % default value: 6pt
\begin{tabular}{llccccccccc}\toprule
& &\textbf{BBBP} &\textbf{BACE} &\textbf{Tox21} &\textbf{ToxCast} &\textbf{SIDER} 
&\textbf{ClinTox} &\textbf{MUV} &\textbf{HIV} \\
% \midrule
% No pre-training (baseline)~\citep{Hu*2020Strategies} &65.8 ± 4.5 &70.1 ± 5.4 &74.0 ± 0.8 &63.4 ± 0.6 &57.3 ± 1.6 &58.0 ± 4.4 &71.8 ± 2.5 &75.3 ± 1.9 \\
\midrule
\parbox[t]{2mm}{\multirow{6}{*}{\rotatebox[origin=c]{90}{\textbf{Pre-training}}}} & Self-supervised pre-trained~\citep{Hu*2020Strategies}$\dagger$ &68.8 ± 0.8 &79.9 ± 0.9 &76.7 ± 0.4 &64.2 ± 0.5 &61.0 ± 0.7 &71.8 ± 4.1 &75.8 ± 1.7 &77.3 ± 1.0 \\
& GraphCL pre-trained~\citep{you2020graph} &69.7 ± 0.7 &75.4 ± 1.4 &\textcolor{dark2pink}{73.9 ± 0.7} &\textcolor{dark2pink}{62.4 ± 0.6} &60.5 ± 0.9 &76.0 ± 2.7 &\textcolor{dark2pink}{69.8 ± 2.7} &\textbf{78.5 ± 1.2} \\
& InfoGraph pre-trained~\citep{wang2022evaluating} &66.3 ± 0.6 &\textcolor{dark2pink}{64.8 ± 0.8} &\textcolor{dark2pink}{68.1 ± 0.6} &\textcolor{dark2pink}{58.4 ± 0.6} &\textcolor{dark2pink}{57.1 ± 0.8} &66.3 ± 0.6 &\textcolor{dark2pink}{44.3 ± 0.6} &\textcolor{dark2pink}{70.2 ± 0.6} \\
& JOAOv2 pre-trained~\citep{wang2022evaluating} &66.4 ± 0.9 &\textcolor{dark2pink}{67.4 ± 0.7} &\textcolor{dark2pink}{68.2 ± 0.8} &\textcolor{dark2pink}{57.0 ± 0.5} &59.1 ± 0.7 &64.5 ± 0.9 &\textcolor{dark2pink}{47.4 ± 0.8} &\textcolor{dark2pink}{68.4 ± 0.5} \\
& GraphMAE~\citep{hou2022graphmae} &72.0 ± 0.6 &83.1 ± 0.9 &75.5 ± 0.6 &64.1 ± 0.3 &60.3 ± 1.1 &\textbf{82.3 ± 1.2} &76.3 ± 2.4 &77.2 ± 1.0 \\
& GraphLoG~\citep{xu2021self} &\textbf{72.5 ± 0.8} &\textbf{83.5 ± 1.2} &75.7 ± 0.5 &63.5 ± 0.7 &61.2 ± 1.1 &76.7 ± 3.3 & 76.0 ± 1.1 & 77.8 ± 0.8 \\
\midrule
\parbox[t]{2mm}{\multirow{6}{*}{\rotatebox[origin=c]{90}{\textbf{No pre-training}}}} & No augmentation (baseline)~\citep{Hu*2020Strategies} &65.8 ± 4.5 &70.1 ± 5.4 &74.0 ± 0.8 &63.4 ± 0.6 &57.3 ± 1.6 &58.0 ± 4.4 &71.8 ± 2.5 &75.3 ± 1.9 \\
&GraphLoG augmented &\textcolor{dark2pink}{65.6 ± 1.0} &82.5 ± 1.2 &\textcolor{dark2pink}{73.2 ± 0.5} &63.6 ± 0.4 &60.9 ± 0.7 &72.5 ± 3.5 &72.4 ± 1.5 &\textcolor{dark2pink}{74.4 ± 1.5} \\
&LapPE augmented &67.1 ± 1.6 &80.4 ± 1.5 &76.6 ± 0.3 &65.9 ± 0.7 &59.3 ± 1.7 &76.4 ± 2.3 &75.6 ± 0.8 &75.6 ± 1.1 \\
&RWSE augmented &67.0 ± 1.4 &79.6 ± 2.8 &76.3 ± 0.5 &65.6 ± 0.3 &58.5 ± 1.4 &74.5 ± 4.4 &75.0 + 1.0 &78.1 ± 1.5 \\
&AllPSE augmented &67.6 ± 1.2 &77.0 ± 4.4 &75.9 ± 1.0 &63.9 ± 0.3 &\textbf{63.0 ± 0.6} &72.6 ± 4.3 &\textcolor{dark2pink}{67.9 ± 0.7} &75.4 ± 1.5 \\
&\textbf{\method} augmented &66.2 ± 0.9 &80.8 ± 3.1 &\textbf{77.4 ± 0.8} &\textbf{66.3 ± 0.8} &61.1 ± 1.6 &78.8 ± 3.8 &\textbf{76.6 ± 1.2} &77.2 ± 1.5 \\
%GPSE augmented (MNIST) &67.5 ± 1.3 &79.0 ± 1.4 &76.2 ± 0.7 &\textbf{66.0 ± 0.6} &61.1 ± 0.6\\

\bottomrule
\multicolumn{10}{l}{$\dagger$ Best test performance reported out of four pre-training strategies (see Table~\ref{tab:molnet_full} for expanded results).}
\vspace{-9pt}
\end{tabular}
\end{table*}

%% file: tables/graph_bench.tex
\begin{table*}[!htp]\centering
\caption{OOD transferability of MolPCBA-trained GPSE to datasets that vary in graph size and connectivity patterns. Dataset information and statistics are available in Appendix \ref{appendix:datasets}.}\label{tab:graphbench}
\footnotesize
%\scriptsize
% \hspace*{-0.12cm}
\begin{tabular}{l|cc|cc}\toprule
&\textbf{Peptides-struct} &\textbf{Peptides-func} &\textbf{CIFAR10} &\textbf{MNIST} \\
&\textbf{MAE} $\downarrow$ &\textbf{AP} $\uparrow$ &\textbf{ACC} (\%) $\uparrow$ &\textbf{ACC} (\%) $\uparrow$ \\
Avg. \# nodes       &150.9  &150.9  &117.6  &70.6 \\
Avg. connectivity   &0.022  &0.022  &0.069  &0.117 \\
\midrule
% GCN &-- &-- &55.255 ± 1.527 &96.485 ± 0.252 \\
GIN &-- &-- &55.26 ± 1.53 &96.49 ± 0.25 \\
GINE &0.3547 ± 0.0045 &0.5498 ± 0.0079 &-- &-- \\
GatedGCN~\citep{bresson2017residual} &0.3420 ± 0.0013 &0.5864 ± 0.0077 &67.31 ± 0.31 &97.34 ± 0.14 \\
Graph MLP-Mixer~\citep{he2022generalization} &\underline{0.2475 ± 0.0020} &\underline{0.6920 ± 0.0054} &\textbf{72.46 ± 0.36} &\textbf{98.35 ± 0.10} \\
DRew+GCN~\citep{gutteridge2023drew} & 0.2781 ± 0.0028 &\textbf{0.6996 ± 0.0076} &-- &-- \\
PR-MPNN~\citep{qian2023probabilistically}  &0.2477 ± 0.0005 &0.6825 ± 0.0086 &-- &-- \\
\midrule
GCN+LapPE~\citep{tönshoff2023did} &0.2492 ± 0.0019 &0.6218 ± 0.0055 &-- &-- \\
GCN+RWSE &0.2574 ± 0.0020 &0.6067 ± 0.0069 &-- &-- \\
GCN+\textbf{\method} &0.2487 ± 0.0011 &0.6316 ± 0.0085 &-- &-- \\
\midrule
GPS+none &0.3817 ± 0.0207 &0.6231 ± 0.0252 &71.67 ± 0.01 &98.05 ± 0.00 \\
GPS+(RWSE/LapPE)~\citep{rampavsek2022recipe} &0.2500 ± 0.0005 &0.6535 ± 0.0041 &72.30 ± 0.36 &98.05 ± 0.13 \\
% GatedGCN+GPSE &0.xxxx ± 0.xxxx &0.xxxx ± 0.xxxx &xx.xxx ± xx.xxx &xx.xxx ± xx.xxx \\
% GINE+GPSE &0.xxxx ± 0.xxxx &0.xxxx ± 0.xxxx &xx.xxx ± xx.xxx &xx.xxx ± xx.xxx \\
GPS+AllPSE	&0.2509 ± 0.0028 &0.6397 ± 0.0092 &72.05 ± 0.35 &98.08 ± 0.12 \\
GPS+\textbf{\method} &\textbf{0.2464 ± 0.0025} &0.6688 ± 0.0151 &\underline{72.31 ± 0.25} &\underline{98.08 ± 0.13}\\
\bottomrule
\end{tabular}
%\vspace{-10pt}
\end{table*}

%% file: tables/node_bench_new.tex
\begin{table}[htp]
\captionsetup{width=.5\linewidth}
\centering
\caption{OOD transferability to OGB node classification benchmarks. Best model in each model-dataset category is \underline{underlined}, best overall model for each dataset is indicated in \textbf{bold}.}\label{tab:nodebench_full}
% \scriptsize
\vspace{2pt}
\footnotesize
\setlength\tabcolsep{5pt} % default value: 6pt
\begin{tabular}{lcccc}\toprule
&\textbf{arXiv} &\textbf{Proteins} \\
&\textbf{ACC} (\%) $\uparrow$ &\textbf{AUROC} (\%) $\uparrow$ \\
\midrule
GCN+none     &71.62 ± 0.23 &\underline{80.44 ± 0.56} \\
GCN+LapPE    &70.89 ± 0.20 &80.38 ± 0.16 \\
GCN+\textbf{\method}     &\underline{71.70 ± 0.26} &80.25 ± 0.19 \\
\midrule
SAGE+none    &\underline{\textbf{72.36 ± 0.43}} &\underline{80.35 ± 0.07} \\
SAGE+LapPE   &71.63 ± 0.16 &80.27 ± 0.45 \\
SAGE+\textbf{\method}    &\underline{\textbf{72.34 ± 0.19}} &80.14 ± 0.22 \\
\midrule
% GIN(E)                  &\xmark &71.82 ± 0.17 &83.70 ± 0.27 \\
%                         &\cmark &71.85 ± 0.09 &83.60 ± 0.09 \\
% \midrule
GAT(E)v2+none    &71.69 ± 0.21 &83.47 ± 0.13 \\
GAT(E)v2+LapPE   &71.30 ± 0.27 & 83.25 ± 0.05 \\
GAT(E)v2+\textbf{\method}    &\underline{72.17 ± 0.42} &\underline{\textbf{83.51 ± 0.11}} \\
\midrule
Transformer+none &57.00 ± 0.79 &73.93 ± 1.44 \\
Transformer+LapPE&57.21 ± 0.25 &74.05 ± 0.11 \\
Transformer+\textbf{\method} &\underline{59.17 ± 0.21} &\underline{74.67 ± 0.74} \\
\midrule
GPS+none    &70.60 ± 0.28 &69.55 ± 5.67 \\
GPS+LapPE   &70.62 ± 0.41 & 70.80 ± 4.14 \\
GPS+\textbf{\method}    &\underline{70.89 ± 0.36} &\underline{72.05 ± 3.75} \\

%\midrule
%\textit{Best}-none  &71.74 ± 0.29 &83.47 ± 0.13 \\
%\textit{Best}-LapPE  &71.63 ± 0.16 &83.25 ± 0.05 \\
%\textit{Best}-GPSE  &\textbf{72.19 ± 0.32} &\textbf{83.51 ± 0.11} \\
\bottomrule
\end{tabular}
\vspace{-0.1in}
\end{table}

%% file: tables/synthetic.tex
\begin{table}[!htp]\centering
% \begin{table}[!htp]\centering
% \vspace{-0.12in}
% \vspace{-13pt}
% \vspace{-5pt}
\caption{Synthetic graph benchmarks with ten times stratified five-fold CV evaluated on ACC (\%) $\uparrow$.}\label{tab:synth}
\vspace{2pt}
\fontsize{8.6}{8.6}\selectfont
\setlength\tabcolsep{4pt} % default value: 6pt
% \footnotesize
% \scriptsize
% \tiny
% \setlength\tabcolsep{5pt} % default value: 6pt
% \begin{tabular}{lcccc}\toprule
\begin{tabular}{@{\extracolsep{2pt}}lcccc@{}}\toprule
% &CSL &EXP \\
&\multicolumn{2}{c}{\textbf{CSL}} &\multicolumn{2}{c}{\textbf{EXP}}\\
\cmidrule{2-3}
\cmidrule{4-5}
&Train &Test &Train &Test\\
\midrule
GIN       &10.0 ± 0.0   &10.0 ± 0.0     &49.8 ± 1.8    &48.7 ± 2.2 \\
GIN+rand  &11.6 ± 3.7   &12.7 ± 6.4     &51.0 ± 2.0    &51.3 ± 2.9 \\
GIN+\textbf{\method}  &98.2 ± 1.5   &42.9 ± 7.9     &84.6 ± 6.8    &68.3 ± 7.5 \\
\midrule
GIN+LapPE &100.0 ± 0.0  &92.5 ± 4.2     &99.9 ± 0.2    &99.5 ± 0.8 \\
GIN+RWSE  &100.0 ± 0.0  &100.0 ± 0.0    &99.7 ± 0.2    &99.7 ± 0.6 \\
\bottomrule
\end{tabular}
% \vspace{-0.05in}
% \vspace{-5pt}
\end{table}

%% file: appendix.tex
\section{Positional and structural encoding tasks}\label{sec:posenc}

The success of \method relies on reliably learning diverse positional and structural encodings (PSE) for graphs during training. Here, we elaborate on the formulations of the PSEs introduced in \S\ref{sec:target_pse}.

We consider a simple undirected and unweighted graph $G = (V, E)$ as a tuple of the vertex set $V$ and the edge set $E$, with no node or edge features. We denote the number of nodes and the number of edges as $n = |V|$ and $m = |E|$, respectively. Then, the corresponding adjacency matrix representing the graph $G$ is a symmetric matrix $\mathbf{M} \in \{0, 1\}^{n \times n}$, where $\mathbf{M}_{ij} = 1$ if $(v_i, v_j) \in E$ and 0 otherwise. The graph Laplacian $\mathbf{L}$ is defined as

\begin{equation}
    \mathbf{L} = \mathbf{D} - \mathbf{M}
\end{equation}

where $\mathbf{D} \in \mathbb{N}^{n \times n}$ is a diagonal matrix whose entries correspond to the degree of a vertex in the graph, $\mathbf{D}_{ii} = deg(v_i) = |\mathcal{N}(v_i)| = |\{ u | (v_i, u) \in E\}|$.

The graph Laplacian is a real symmetric matrix, thus having a full eigendecomposition as

\begin{equation}
    \mathbf{L} = \mathbf{U} \mathbf{\Lambda} \mathbf{U}^\top
\end{equation}

where, $\mathbf{\Lambda}_{ii} = \lambda_i$ and $\mathbf{U}_{[:, i]} = u_i$ are the $i^{\text{th}}$ eigenvalue and eigenvector (an eigenpair) of the graph Laplacian.
We follow the convention of indexing the eigenpair from the smallest to the largest eigenvalue, i.e., $0 = \lambda_1 \leq \lambda_2 \leq \dots \leq \lambda_n$. We further denote $\hat{\mathbf{U}}$ (and analogously the subdiagonal matrix $\hat{\mathbf{\Lambda}}$) as the matrix of Laplacian eigenvectors corresponding to non-trivial eigenvalues.

\begin{equation}
    \hat{\mathbf{U}} = \mathbf{U}_{[:, \{i | \lambda_{i} \neq 0\}]}
\end{equation}

Finally, we denote the ($\ell_2$) normalization operation as $normalize(x) := \frac{x}{\|x\|_2}$

\paragraph{Laplacian eigenvector positional encodings (LapPE)}

LapPE is defined as the absolute value of the $\ell_2$ normalized eigenvectors associated with non-trivial eigenvalues. We use the first four LapPE to train \method by default.

\begin{equation}
    \text{LapPE}_{i} = |normalize(\hat{\mathbf{U}}_{[:, i]})|
\end{equation}

The absolute value operation is needed to counter the sign ambiguity of the graph Laplacian eigenvectors, a known issue to many previous works that use the Laplacian eigenvectors to augment the models~\cite{benchmarking_gnns,lim2022sign}. However, common strategies to overcome the sign ambiguity issue such as random sign flipping~\cite{benchmarking_gnns}  or constructing sign invariant function~\cite{lim2022sign} do not resolve our issue here as we are trying to \textit{recover} the PEs rather than using them as features.

We have, however, conducted an ablation study to find if better strategies may exist than taking the absolute value of the eigenvectors. new set of experiments to demonstrate this aspect further. In this study, we pretrained four different versions of \method using 5\% MolPCBA-subset: 
\begin{enumerate}
    \item \emph{GPSE-abs} takes the absolute value of the LapPE (default setting in our paper)
    \item \emph{GPSE-noabs} does not take the absolute value of the LapPE,
    \item \emph{GPSE-signinvar} uses a sign invariant loss function for LapPE by taking the minimum of the losses from both signs,
    \item \emph{GPSE-SignNet} uses a randomly initialized SignNet \citep{lim2022sign} model to generate sign invariant features as the training target for GPSE.
\end{enumerate}

Our results in Table~\ref{tab:abl_gpse_lap} indicate that using the default absolute handling of LapPE results in similar or better performance than other strategies, indicating the effectiveness of using the absolute LapPE for training GPSE. 
Nevertheless, investigating better strategies for learning \textit{invariant} representations for eigenvectors is an interesting venue for future studies.

We additionally use the eigenvalues as a graph-level regression task for training \method.

\paragraph{Electrostatic potential positional encodings (ElstaticPE)}

The pseudoinverse of the graph Laplacian $\mathbf{L}^\dagger$ has a physical interpretation that closely relates to the electrostatic potential between two nodes in the graph $G$ when each node is treated as a charged particle ~\cite{kreuzer2021rethinking} and can be computed as

\begin{equation}
    \mathbf{L}^\dagger = \mathbf{U} \mathbf{\Lambda}^\dagger \mathbf{U}^\top = \hat{\mathbf{U}} \hat{\mathbf{\Lambda}}^{-1} \hat{\mathbf{U}}^\top
\end{equation}

We further subtract each column of $\mathbf{L}^\dagger$ by its diagonal value to set zero ground state such that each node's potential on itself is 0.
% \todo{Dominique, please check if this makes sense}.

\begin{equation}
    \mathbf{Q} = \mathbf{L}^\dagger - diag(\mathbf{L}^\dagger) \mathbf{1}_{n}
\end{equation}

The final ElstaticPE is a collection of aggregated values for each node, that summarizes the electrostatic interaction of a node with all other nodes:
\begin{enumerate}
    \item Minimum potential from $v_i$ to $v_j$: $\text{ElstaticPE}_1(i) = \text{min}(\mathbf{Q}_{[:, i]})$
    \item Average potential from $v_i$ to $v_j$: $\text{ElstaticPE}_2(i) = \text{mean}(\mathbf{Q}_{[:, i]})$
    \item Standard deviation of potential from $v_i$ to $v_j$: $\text{ElstaticPE}_3(i) = \text{std}(\mathbf{Q}_{[:, i]})$
    \item Minimum potential from $v_j$ to $v_i$: $\text{ElstaticPE}_4(i) = \text{min}(\mathbf{Q}_{[i, :]})$
    \item Standard deviation of potential from $v_j$ to $v_i$: $\text{ElstaticPE}_5(i) = \text{std}(\mathbf{Q}_{[i, :]})$
    \item Average interaction on direct neighbors: $\text{ElstaticPE}_6(i) = \text{mean}\left((\mathbf{M}\mathbf{Q})_{[:, i]}\right)$
    \item Average interaction from direct neighbors: $\text{ElstaticPE}_7(i) = \text{mean}\left((\mathbf{M}\mathbf{Q})_{[i, :]}\right)$
\end{enumerate}

% \todo{show the row sum of $\mathbf{Q}$ is a constant, i.e., $\mathbf{Q1} = c\mathbf{1}$, hence we remove from the ElstaticPE collection (note the postprocessing of standardization)}

\paragraph{Random walk structural encodings (RWSE)}

Define the random walk matrix as the row-normalized adjacency matrix $\mathbf{P} := \mathbf{D}^{-1} \mathbf{M}$. Then $\mathbf{P}_{i,j}$ corresponds to the one-step transition probability from $v_i$ to $v_j$.

The $k^\text{th}$ RWSE~\cite{lpse} is defined as the probability of returning back to the starting state of a random walk after exactly $k$ step of random walks:

\begin{equation}
    \text{RWSE}_{k} = diag(\mathbf{P}^{k})
\end{equation}

\paragraph{Heat kernel diagonal structural encodings (HKdiagSE)}

\begin{equation}
    \text{HKdiagSE}_{k} = \sum_{i: \lambda_i \neq 0} e^{-k \lambda_i} normalize(\mathbf{U}_{[:, i]})^2
\end{equation}

\paragraph{Cycle counting structural encodings (CycleSE)}

CycleSE encodes global structural information of the graph by counting the number of $k$-cycles in the graph. For example, a 2-cycle corresponds to an undirected edge, and a 3-cycle corresponds to a triangle.

\begin{equation}
    \text{CycleSE}_{k} = |\{\text{Cycles of length k}\}|
\end{equation}

CycleSE is used as a graph-level regression task for training \method.

% \todo{describe alg}

\paragraph{Normalizing PSEs tasks}

Finally, we perform graph-wide normalization preprocessing step on each node-level PSE task so that they have zero mean and unit standard deviation. This normalization step ensures all PSE targets are on the same scale, making the training process more stable.

% \begin{itemize}
%     \item PE type specific preprocessing
%     \begin{itemize}
%         \item EigVec: abs and l2 norm
%         \item RWSE: ...
%     \end{itemize}
%     \item Standardization (PyG graphnorm -> GNNBench) -> better conditioned prediction targets
% \end{itemize}

\newpage
\section{Implementation details}\label{sec:impl}

% - loss function cosim + mae (why?)
% - skipping virtual node
% - low variance
% - graph wide r2 comp for node-level regression
% - graph-level task treated as one "graph"

% \color{blue}
\subsection{\method computation}
\label{appendix:gpse_comp}
The \method model is built using a GatedGCN backbone~\citep{bresson2017residual} with PSE-specific MLP decoding heads. \method uses random noise drawn from a 20-dimensional standard Gaussian as the input node features. The random features are then projected to the match the hidden dimension, $d$, of the model, resulting in the hidden representations of the first layer:
\begin{equation}
    h_i^{(0)} = \text{ReLU}\Big( x_i W_\text{inp} \Big)
\end{equation}
where $h_i^{(0)} \in \mathbb{R}^{1 \times d}$ indicates the hidden feature of node $i$ in the first layer, $W_\text{inp} \in \mathbb{R}^{20 \times d}$ is the linear projection layer, and $x_i \sim \mathcal{N}(\mathbf{0}, \mathbf{I}) \in \mathbb{R}^{1 \times 20}$ is the random noise. Next, the model enters $L$ layers of GatedGCN convolution layers, where each layer is defined as:
\begin{equation}
    h_i^{(l+1)} = \text{ReLU} \bigg(
    h_i^{(l)} W_1^{(l)}
    + \sum_{j \in \mathcal{N}(i)} \sigma \Big( h_i^{(l)} W_2^{(l)} + h_j^{(l)} W_3^{(l)} \Big) \odot \Big( h_j^{(l)} W_4^{(l)} \Big)
    \bigg)
\end{equation}
where $W_1^{(l)}, W_2^{(l)}, W_3^{(l)}, W_4^{(l)} \in \mathbb{R}^{d \times d}$ are learnable parameters for layer $l$, $\sigma$ is the sigmoid function, and $\odot$ is the elementwise multiplication operator. Finally, the processed hidden feature $h_i^{(L)}$ is decoded via a two-layer MLP to predict the $k^\text{th}$ node-level PSEs, such as LapPE and RWSE.
\begin{equation}
    \hat{y}_{i,k} = \text{ReLU}\Big( h_i^{(L)} W_{k,1} \Big) W_{k,2}
\end{equation}
where $W_{k,1} \in \mathbb{R}^{d \times d}$ and $W_{k,2} \in \mathbb{R}^{d \times 1}$ are learnable parameters for projecting the final hidden representation to the PSE prediction. For graph-level PSEs, such as CycleSE, we use sum-pooling to reduce the hidden representations to graph-level first, and similarly apply a two-layer MLP afterwards. Once trained, we apply \method to extract $h^{(L)}$ for the graphs in the downstream dataset and use it in-place of the traditional PSEs. We set $L$ to $20$, and $d$ to $512$ for our final \method architecture. We also present an ablation study on various architectural choices to demonstrate the effectiveness of our final model setting (Table~\ref{tab:abl_pse_perf}).
\color{black}

\subsection{\method training loss function}
\label{appendix:loss}

We use a combination of $\ell_1$ loss and cosine similarity loss for training \method using the PSE self-supervision defined in Appendix~\ref{sec:posenc}. More specifically, given $M$ number of graphs, and $K$ number of target PSE tasks, we compute the loss as follows:

\begin{equation}
    \mathcal{L} = \sum_{k=1}^{K} \sum_{i=1}^{M}
    \Bigg[
    \bigg( \sum_{j=1}^{|V(G_i)|} \Big| y_{j,k}^{(i)} - \hat{y}_{j,k}^{(i)} \Big| \bigg)
    + \bigg( 1 - \sum_{j=1}^{|V(G_i)|} \tilde{y}_{j,k}^{(i)} \tilde{\hat{y}}_{j,k}^{(i)} \bigg)
    \Bigg]
\end{equation}

where $y_{j,k}^{(i)}$, $\hat{y}_{j,k}^{(i)}$ are the true and predicted values of the $j^\text{th}$ node of $i^\text{th}$ graph for the $k^\text{th}$ PSE task. $\tilde{y}$ and $\tilde{\hat{y}}$ are the $\ell_2$ normalized version of $y$ and $\hat{y}$, respectively. Note that in practice, we compute the loss over mini-batches of graphs rather than over all of the training graphs.

\subsection{Compute environment and resources}

Our codebase is based on GraphGPS~\cite{rampavsek2022recipe}, which uses PyG and its GraphGym module~\cite{pyg,you2020design}. All experiments are run using Tesla V100 GPUs (32GB), with varying numbers of CPUs from 4 to 8 and up to 48GB of memory (except for two cases: (i) 80GB of memory is needed when performing downstream evaluation on MolPCBA, and (ii) 128GB is needed when pre-training \method on the ChEMBL dataset).

We recorded the run time for both the \method pre-computation and the downstream evaluation training loop using Python's \texttt{time.perf\_counter()} function and reported them in Table~\ref{tab:hp_mol_bench},~\ref{tab:hp_transfer_bench},~\ref{tab:hp_node_bench}. We did not report the \method pre-computation time for other downstream benchmarks since they are all within five minutes.

\subsection{Hyperparameters}

% \subsubsection{\method pre-training} \todo{TEXT TEXT TEXT}

\subsubsection{Downstream tasks}

In most of the downstream task hyperparameter searches, we followed the best settings from previous studies~\cite {rampavsek2022recipe}, and primarily tuned the \method encoding parameters, including the \method processing encoder type, the encoded dimensions, the input and output dropout rate of the processing encoder, and the application of batch normalization to the input \method encodings. For completeness, we list all hyperparameters for our main benchmarking studies in Tables~\ref{tab:hp_mol_bench} and~\ref{tab:hp_transfer_bench}.

\input{tables/hp_mol_bench}

\input{tables/hp_transfer_bench}

\input{tables/hp_node_bench}

\textbf{MoleculeNet small benchmarks settings} We used the default GINE architecture following previous studies~\cite{Hu*2020Strategies}, which has five hidden layers and 300 hidden dimensions. For all five benchmarks, we use the same \method processing encoder settings as shown in Table~\ref{tab:hp_molnet_small_bench}.

\textbf{CSL \& EXP synthetic graph benchmarks settings} We follow~\citet{he2022generalization} and use GIN~\cite{xu2018powerful} as the underlying MPNN model, with five hidden layers and 128 dimensions. We use the same \method processing encoder settings for both CSL and EXP as shown in Table~\ref{tab:hp_wl_bench}.

\input{tables/hp_molnet_and_wl}

\clearpage
\section{Theory details}\label{appendix:theory_details}

Message-passing GNNs have receptive fields that grow exponentially with the number of layers. Given two nodes, the influence of one onto the other might become too weak over long graph distances, hindering the learning task. This phenomenon has been referred to as \emph{over-squashing}~\cite{alon2021on}. A similar problem also occurs as the number of layers increases, where the nodes' hidden representations become increasingly similar as the number of layers increase, also known as \emph{over-smoothing}~\cite{li2019deepgcns}.

% \color{blue}
\subsection{Relevance to \method}\label{appendix:theory_details_oo_gpse}

The over-smoothing and over-squashing problems are essential to overcome to effectively learn the positional and structural encodings, especially for those that require \textit{global views of the graph}. For example, the Laplacian eigenvector corresponding to the first non-trivial eigenvalue, also known as the Fiedler vector, corresponds to the solution of the graph min-max cut problem~\cite{ding2001min}. Intuitively, this problem requires accessing the global view of the entire graph as it, colloquially, aims to partition the entire graph into two parts with minimal connections.

A straightforward solution to incorporating more global information into the model is by \textit{stacking more message-passing layers} to increase the receptive field and thus effectively expose the model to information beyond the local structure. However, simply stacking more message-passing layers easily leads to \textit{over-smoothing}, where the messages of each node become increasingly uniform as the number of layers increases. Our usage of the gating mechanism, along with residual connections, effectively mitigates this issue while still exposing the model to more non-local information.

Meanwhile, the model may still have difficulty incorporating global information, even after fixing over-smoothing and stacking more layers, due to \textit{over-squashing}. Informally, over-squashing can be understood as the difficulty in losslessly sending messages between two nodes across the network. This difficulty is primarily because there are only a few possible routes between the two nodes compared to all other available routes to each of the nodes. We mitigate this problem using a \textit{virtual node} that serves as the global information exchange hub to enable global information exchange, bypassing the “few routes” limitation.
\color{black}

\subsection{Formal analysis}
\begin{definition}[Over-squashing]\label{def:oversquashing}
The squashing of a GNN is measured by the influence of one node on the features of another which we interpret as the partial derivative
\[
\frac{\partial h_i^{(r+1)}}{\partial x_j}
\]
for $h_i^{(r)}(x_1,...,x_n)$ the $r$-th hidden feature at node $i$, and $x_j$ the input feature at node $j$. If this quantity converges to 0 as $r$ increases, then the network is said to suffer from \emph{over-squashing}.
\end{definition}

Another common problem with MPNNs is known as \emph{over-smoothing}. It has often been observed that MPNNs with many layers produce node features that are very close or even identical, which limits expressivity and prevents learning. This stems from message-passing being equivalent to a local smoothing operation; too many smoothing iterations result in all nodes converging to identical states.
\begin{definition}[Over-smoothing]\label{def:oversmoothing}
The smoothing of a network can be measured by the norm (for example the $\ell_1$-norm) of the state difference between neighbors, i.e.
\[
\sum_{(i,j) \in E} |h^{(r)}_i - h^{(r)}_j|
\]
where the sum is taken over the edges of the graph. If this quantity converges to 0 as $r$ increases, the network is said to suffer from \emph{over-smoothing}.
\end{definition}

In the following section, we will refer to the relationships between over-squashing and over-smoothing with graph curvature. There are many versions of graph curvature \cite{formanBochnerMethodCell2003, OLLIVIER2009810,sreejithFormanCurvatureComplex2016,topping2022understanding}, all closely related. Here we will only consider the balanced Forman curvature from~\citet{topping2022understanding}.

\begin{definition}[Graph curvature]\label{def:curvature}
For any edge $(i, j)$ in a simple, unweighted graph $G$, its contribution to graph curvature is given by
\[
\textup{Ric}(i,j) = \frac{2}{d_i}+\frac{2}{d_j}-2 + |{\#_\Delta (i,j) }| \left(\frac{2}{\textup{max}\{d_i,d_j\}} + \frac{1}{\textup{min}\{d_i,d_j\}} \right)+\frac{\gamma_{\textup{max}}}{\textup{max}\{d_i,d_j\}} \left(|{\#_{\square}^i}|+|{\#^j_\square}|\right)
\]
where $\#_\square^i$ is the number of 4-cycles containing the node $i$ (diagonals not allowed), $\#_\Delta^i$ is the number of 3-cycles containing $i$, $d_i$ is the degree of $i$ and $\gamma_{\text{max}}$ is the maximum over nodes $k$ of the number of 4-cycles that pass through the nodes $i,j$ and $k$.
\end{definition}

It can then be shown that negative curvature causes over squashing \cite{topping2022understanding, nguyenRevisitingOversmoothingOversquashing2023} and positive curvature causes over smoothing \cite{nguyenRevisitingOversmoothingOversquashing2023, giraldoUnderstandingRelationshipOversmoothing2022}.

Next, we show that rewiring the graph by adding a virtual node increases the balanced Forman curvature of the graph at most edges.
\begin{proposition}\label{prop:VNincreasesCurvature}
    The balanced Forman Curvature is increased for most edges when adding a virtual node such that
    \[\textup{Ric}(i,j) - \textup{Ric}^{+\textup{VN}}(i,j) \leq \frac{1}{(d_i-\delta)^2+d_i-\delta}-\frac{2\delta}{d_i^2+d_i},\]
    where $d_i$ is the degree of the most connected node of the edge $(i,j)$ and $\delta = d_i-d_j$.
\end{proposition}
\textbf{Proof.}
$\#_\square$ is invariant when adding virtual node because it automatically creates diagonals in the new 4-cycles. Therefore, $\gamma_{\text{max}}$ is also invariant. As for $d_i$, $d_j$ and $\#_\Delta$, they are all increased by 1:
$$\text{Ric}(i,j)-\text{Ric}^+(i,j) \approx 2\left(\frac{1}{d_i}+\frac{1}{d_j}-\frac{1}{d_i+1}-\frac{1}{d_j+1}\right) + |\#_\Delta (i,j) |\left(\frac{2}{\text{max}\{d_i,d_j\}} + \frac{1}{\text{min}\{d_i,d_j\}}\right) $$
$$\hspace{2cm} - \left(|\#_\Delta (i,j) |+1)(\frac{2}{\text{max}\{d_i,d_j\}+1} + \frac{1}{\text{min}\{d_i,d_j\}+1}\right)$$
We can let $d_i \geq d_j$ without loss of generality. The inequality is not influenced by the introduction of a virtual node:

\[\text{Ric}(i,j)-\text{Ric}^+(i,j) \approx 2\left(\frac{1}{d_i^2+d_i}+\frac{1}{d_j^2+d_j}\right) + |\#_\Delta (i,j)|\left(\frac{2}{d_i} + \frac{1}{d_j}\right) - (|\#_\Delta (i,j)|+1)\left(\frac{2}{d_i+1} + \frac{1}{d_j+1}\right)\]
$$\text{Ric}(i,j)-\text{Ric}^+(i,j) \approx 2\left(\frac{1}{d_i^2+d_i}+\frac{1}{d_j^2+d_j}\right) + |\#_\Delta (i,j) |\left(\frac{2}{d_i^2+d_i} + \frac{1}{d_j^2+d_j}\right) - \left(\frac{2}{d_i+1} + \frac{1}{d_j+1}\right)$$
The number of triangles is upper bounded by the least connected node's degree minus 1,
$|\#_\Delta| \leq d_j -1$. We then have:

\begin{align*}
    \text{Ric}(i,j)-\text{Ric}^+(i,j) &\leq 2\left(\frac{1}{d_i^2+d_i}+\frac{1}{d_j^2+d_j}\right) + (d_j-1)\left(\frac{2}{d_i^2+d_i} + \frac{1}{d_j^2+d_j}\right) - \left(\frac{2}{d_i+1} + \frac{1}{d_j+1}\right)\\
     &= -\frac{2(d_i-1)}{d_i^2+d_i}-\frac{d_j-2}{d_j^2+d_j} + (d_j-1)\left(\frac{2}{d_i^2+d_i} + \frac{1}{d_j^2+d_j}\right)\\
     &= -\frac{2(d_i-1-d_j+1)}{d_i^2+d_i}-\frac{d_j-2-d_j+1}{d_j^2+d_j}\\
     &= \frac{1}{d_j^2+d_j} -\frac{2(d_i-d_j)}{d_i^2+d_i}
\end{align*}
Let's call the difference between the two nodes' degrees $\delta$. We get:
$$\text{Ric}(i,j)-\text{Ric}^+(i,j) \leq \frac{1}{(d_i-\delta)^2+d_i-\delta}-\frac{2\delta}{d_i^2+d_i} $$
This upper bound gives us a good insight on the general behavior of the curvature when adding a virtual node.

\textbf{First case:} 
The upper bound of the difference is negative for $\delta \neq 0$ and $d_j\neq 1$. This means that for the most cases, the addition of the virtual node clearly increases the curvature.

\textbf{Second case:}
The upper bound is positive for $d_j =1$ (ie. $d_i-\delta =1$). However, the isolated edges are not responsible for bottleneckness. It is to be noted that such \textit{isolated }edges never have negative curvature, neither before nor after the addition of the virtual node. This is a direct consequence of the curvature definition.

\textbf{Third case:} The upper bound is positive for $\delta = 0$. This comes as a surprise and might need future work. It is to be noted that the upper bound tends toward 0 pretty quickly as  (as $\frac{1}{d_i^2}$), thus, by the nature of the upper bound, the addition of the virtual node should still increase curvature for most of the cases where $\delta =0$.

Note that we didn't include the 4-cycle term, because this term is inversely proportional to the number of triangles, and is therefore equal to 0 when the number of triangle is maximal. Otherwise, as the number of triangle decreases, the upper bound on the 4-cycle term increases, in a slower fashion. Thus, the upper bound still holds.

\newpage
\section{Datasets}
\label{appendix:datasets}

\input{tables/dataset_info}

\input{tables/dataset_stats}

\subsection{Pre-training datasets}

\textbf{MolPCBA}~\cite{hu2020open} (MIT License) contains 400K small molecules derived from the MoleculeNet benchmark~\cite{wu2018moleculenet}. There are 323,555 unique molecular graphs in this dataset.

\textbf{ZINC}~\cite{gomez2018automatic} (Apache 2.0 License) contains 250K drug-like commercially available small molecules sampled from the full ZINC~\cite{irwin2012zinc} database. There are 219,2384 unique molecular graphs in this dataset.

\textbf{GEOM}~\cite{axelrod2022geom} (CC0 1.0 license) consists of 300K drug-like small molecules. There are 169,925 unique molecular graphs in this dataset.

\textbf{ChEMBL}~\cite{gaulton2012chembl}\footnote{We used release 32 of ChEMBL: http://doi.org/10.6019/CHEMBL.database.32} (CC BY-SA 3.0 License) consists of 1.4M drug-like bioactive small molecules. There are 970,963 unique molecular graphs in this dataset.

\textbf{PCQM4Mv2}~\cite{hu2021ogb} (CC BY 4.0 License) contains 3.4M small molecules from the PubChemQC~\cite{nakata2017pubchemqc} project. The ground-state electronic structures of these molecules were calculated using Density Functional Theory. There are 273,920 unique molecular graphs in this dataset.

\subsubsection{Extracting unique molecular graph structures}

To extract unique molecular graphs, we use RDKit with the following steps:
\begin{enumerate}[topsep=-1pt,itemsep=0pt,partopsep=0pt,parsep=3pt,leftmargin=3em,itemindent=0em]
    \item For each molecule, convert all its heavy atoms to carbon and all its bonds to single-bond.
    \item Convert the modified molecules into a list of SMILES strings.
    \item Reduce the list to unique SMILES strings using the \texttt{set()} operation in Python.
\end{enumerate}

\subsection{Downstream evaluation datasets}

\textbf{ZINC-subset}~\cite{benchmarking_gnns} (Custom license, free to use) is a 12K subset of the ZINC250K dataset~\cite{gomez2018automatic}. Each graph is a molecule whose nodes are atoms (28 possible types) and whose edges are chemical bonds (3 possible types). The goal is to regress the constrained solubility~\cite{benchmarking_gnns} (logP) of the molecules. This dataset comes with a pre-defined split with 10K training, 1K validation, and 1K testing samples.

\textbf{MolHIV \& MolPCBA}~\cite{hu2020open} (MIT License) are molecular property prediction datasets derived from the MoleculeNet benchmarks~\cite{wu2018moleculenet}. Each graph represents a molecule whose nodes are atoms (9-dimensional features containing atom type, chirality, etc.) and whose edges are chemical bonds. The goal for MolHIV is to predict molecules' ability to inhibit HIV virus replication as a binary classification task. On the other hand, MolPCBA consists of 128 binary classification tasks that are derived from high-throughput bioassay measurements. Both datasets come with pre-defined splits based on the \textit{scaffold splitting} procedure~\cite{Hu*2020Strategies}.

\paragraph{PCQM4Mv2-subset}~\cite{hu2021ogb,rampavsek2022recipe} (CC BY 4.0 License) is a subsampled version of PCQM4Mv2~\cite{hu2021ogb} using random 10\% for training, 33\% for validation, and the original validation set for testing. The molecular graphs are processed the same way as for MolHIV and MolPCBA, where each node is an atom, and each edge is a chemical bond. The task is to regress the HOMO-LUMO energy gap (in electronvolt) given the molecular graph. We note that the subsetted splits used in this work could be different from those used in~\citet{rampavsek2022recipe} as the \texttt{numpy} random generator may not be persistent across \texttt{numpy} versions\footnote{\url{https://stackoverflow.com/a/71790820/12519564}}. To enable reproducibility, we also make our split indices available for future studies to benchmark against.
% \todo{make}.

\textbf{MoleculeNet small datasets}~\cite{Hu*2020Strategies} (MIT License) We follow~\citet{sun2022does} and use the selection of five small molecular property prediction datasets from the MoleculeNet benchmarks, including BBBP, BACE, Tox21, ToxCast, and SIDER. Each graph is a molecule, and it is processed the same way as for MolHIV and MolPCBA. All these datasets adopt the \textit{scaffold splitting} strategy that is similarly used on MolHIV and MolPCBA.

\textbf{Peptides-func \& Peptides-struct}~\cite{dwivedi2022long} (CC BY-NC 4.0 License) both contain the same 16K peptide graphs retrieved from SAT-Pdb~\cite{singh2016satpdb}, whose nodes are residues. The two datasets differ in the graph-level tasks associated with them. Peptides-func aims to predict the functions of each peptide (10-way  multilabel classification), while Peptides-struct aims to regress 11 structural properties of each peptide. Splitting is done via meta-class holdout based on the original labels of the peptides.

\textbf{CIFAR10 \& MNIST}~\cite{benchmarking_gnns} (CC BY-SA 3.0 and MIT License) are derived from the CIFAR10 and MNIST image classification benchmarks by converting the images into SLIC superpixel graphs with 8 nearest neighbors for each node (superpixel). The 10-class classification and the splitting follow the original benchmarks (MNIST 55K/5K/10K, CIFAR10 45K/5K/10K train/validation/test splits).

\textbf{CSL}~\cite{benchmarking_gnns} (MIT License) contains 150 graphs that are known as circular skip-link graphs~\cite{murphy2019relational}. The goal is to classify each graph into one of ten isomorphism classes. The dataset is class-balanced, where each isomorphism class contains 15 graph instances. Splitting is done by stratified five-fold cross-validation.

\textbf{EXP}~\cite{abboud2020surprising} (\textit{unknown} license) contains 600 pairs of graphs (1,200 graphs in total) that cannot be distinguished by 1\&2-WL tests. The goal is to classify each graph into one of two isomorphism classes. Splitting is done by stratified five-fold cross-validation.

\textbf{arXiv}~\cite{hu2020open} (ODC-BY License) is a directed citation graph whose nodes are arXiv papers and whose edges are citations. Each node is featured by a 128-dimensional embedding obtained by averaging over the word embeddings of the paper's title and abstract. The goal is to classify the papers (nodes) into one of 40 subject areas of  arXiv CS papers. Papers published before 2017 are used for training, while the remaining papers that are published before and after 2019 are used for validation and testing.

\textbf{Proteins}~\cite{hu2020open} (CC0 License) is an undirected and weighted graph representing the interactions (edges) between proteins (nodes) obtained from eight species. Each edge has eight channels, corresponding to different types of protein interaction evidence. The task is to predict proteins' functions (112-way multilabel classification). The splitting is done by holding out proteins that correspond to specific species.

\newpage
\section{Visualization of GPSE encodings on 1-WL indistinguishable graph pairs}\label{appendix:wl}

\begin{definition}[Graph isomorphism]\label{def:graph_isomorphism}
Two graphs $G$ and $H$ are isomorphic if there exists a bijection $f$ between their vertex sets
$$f: V(G) \rightarrow V(H)$$
s.t. any two vertices $u, v \in G$ are adjacent in $G$ if and only if $f(u), f(v) \in H$ are adjacent in $H$.
\end{definition}

\begin{figure}[!htp]
    \centering
    \begin{minipage}{0.5\textwidth}
        \centering
        \includegraphics[width=0.8\textwidth]{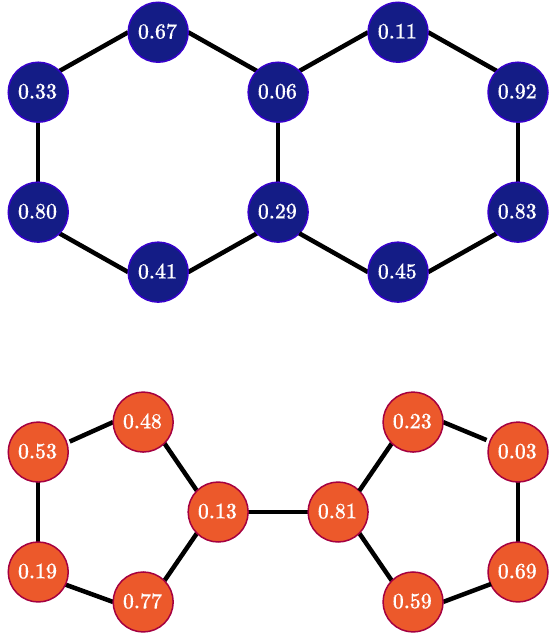} % first figure itself
    \end{minipage}\hfill
    \begin{minipage}{0.5\textwidth}
        \centering
        \includegraphics[width=0.8\textwidth]{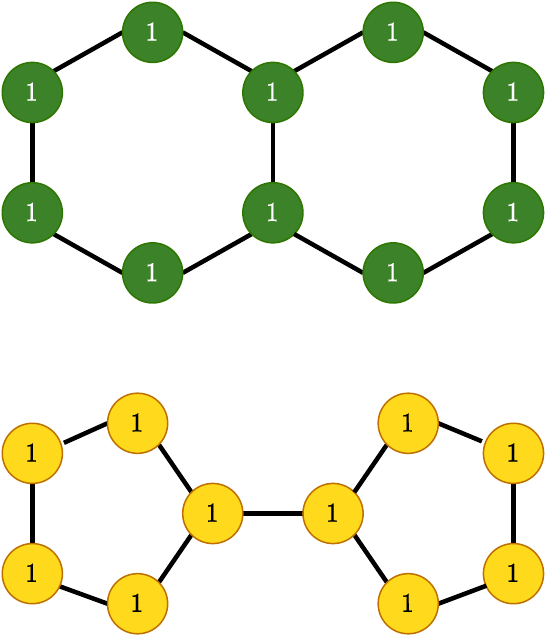} % second figure itself
    \end{minipage}
\end{figure}
\begin{figure}[!htp]
    \centering
    \begin{minipage}{0.5\textwidth}
        \centering
        \includegraphics[width=\textwidth]{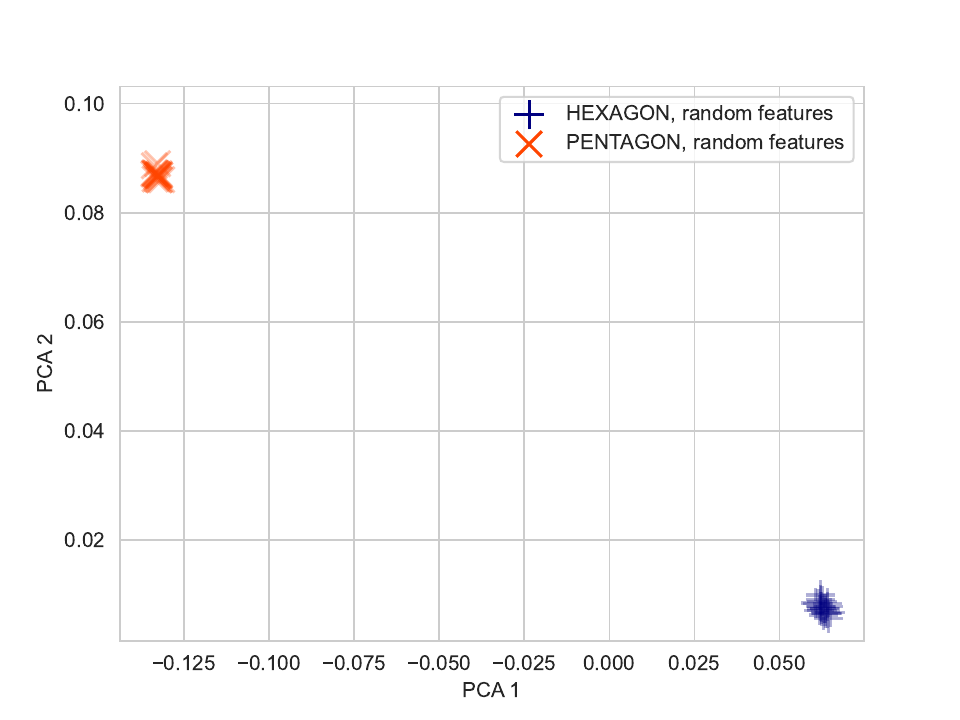} % first figure itself
    \end{minipage}\hfill
    \begin{minipage}{0.5\textwidth}
        \centering
        \includegraphics[width=\textwidth]{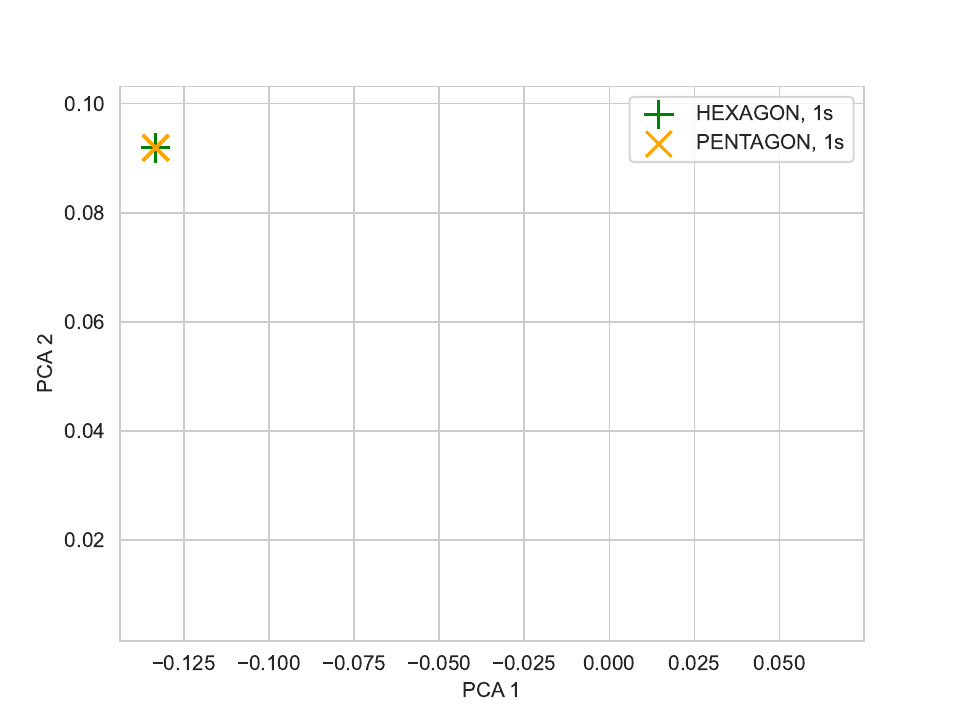} % second figure itself
    \end{minipage}
    \caption{2D PCA visualization of \method encodings on 1-WL indistinguishable graph pairs.
    Applying \method with randomly initialized node features results in distinct encodings for HEXAGON (\textcolor{BlueViolet}{indigo}) and PENTAGON (\textcolor{RedOrange}{orange}) graphs (left). The same graphs cannot be distinguished by our encoder when the node features are set to 1 for each node (right).
    } \label{fig:vis_1wl_graphs}
\end{figure}

The 1-Weisfeiler-Leman (WL) test is an algorithm akin to message-passing that is commonly used to detect non-isomorphic graphs. It can also be viewed as a measure of expressivity: A GNN that can distinguish all pairs of non-isomorphic graphs that can also be distinguished by the 1-WL test is called ``1-WL expressive''.

In \S\ref{sec:expr}, we discussed that our \method is expressive enough to discern graphs that are 1-WL indistinguishable, a well-known limitation of MPNNs \cite{xu2018powerful}. However, \citet{sato2021random} show that the 1-WL expressivity limitation exists only when each node employs identical features; appending random features to the nodes is sufficient to achieve expressivity that goes beyond 1-WL. \method leverages precisely this property by replacing the node features by vectors drawn from a standard normal distribution, such that no two graphs have identical node features.

Here, we demonstrate the importance of these random node features empirically. Consider the following two graphs displayed in Figure~\ref{fig:vis_1wl_graphs}: One resembles two hexagons sharing an edge (referred to as HEXAGON), while the other resembles two pentagons connected by an edge (PENTAGON). These are a well-known pair of non-isomorphic but 1-WL indistinguishable graphs. Non-isomorphism implies that these graphs do not share the same connectivity (see Def.~\ref{def:graph_isomorphism} for a formal definition). The WL test also has its limitations: While two graphs that are deemed non-isomorphic are guaranteed to be so, there are cases where it cannot detect non-isomorphism as in the HEXAGON-PENTAGON case.

% Our experimental pipeline consist of generating two synthetic datasets, both consisting of 20 copies of HEXAGON and PENTAGON graphs each. The two datasets are identical except one has all node features set to 1, while the other has features drawn from a random Normal assigned to each node, thus mirroring the actual \method training pipeline. We then apply \method to both datasets and generate aggregate encodings by averaging the resulting 512-dimensional node encodings for each graph. For visualization purposes, we then apply dimensionality reduction to these graph-level encodings by fitting a 2-dimensional PCA to \method encodings generated on ZINC, and then applying it to the encodings from the synthetic data. 
In our experiment, we create two sets of graphs, both consisting of 20 copies of HEXAGON and PENTAGON graphs each. The two sets are identical except one has all node features set to 1, while the other has features drawn from a random Normal assigned to each node, thus mirroring the actual \method training pipeline. We then apply an already trained \method encoder (trained on ZINC) to both sets and for each graph we generate aggregated (graph-level) encodings by averaging the obtained 512-dimensional node encodings from \method. For visualization purposes, we then apply dimensionality reduction to these graph-level encodings by first fitting a 2-dimensional PCA to \method encodings generated on ZINC, and then applying it to the encodings from the synthetic data.

As shown in Figure~\ref{fig:vis_1wl_graphs}, applying \method to graphs with randomly initialized node features results in distinct encodings for HEXAGON (\textcolor{BlueViolet}{indigo}) and PENTAGON (\textcolor{RedOrange}{orange}) graphs (Fig.~\ref{fig:vis_1wl_graphs}, left). The same graphs cannot be distinguished by our encoder when the node features are 1 for each node (Fig.~\ref{fig:vis_1wl_graphs}, right). The same result is observed when analysing the graph-level PCA embeddings, that can clearly separate the two types of graphs when random node features are used by \method, but not otherwise. This underlines the importance of randomized node features in \method.

\clearpage
\section{\method training and inference times}

\input{tables/gpse_train_time}

\input{tables/gpse_test_time}

\clearpage
\section{\method vs. conventional PSE scaling experiments}
\label{appendix:scaling}

\input{tables/graph_number_scaling}

\begin{figure}[h]
    \centering
    \begin{subfigure}[t]{0.35\textwidth}
        \includegraphics[width=\textwidth]{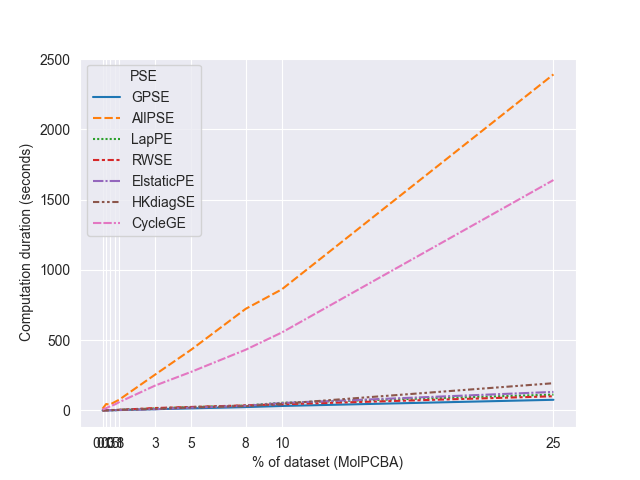}
        \caption{\method + individual PSEs + combined PSEs (AllPSE)}
    \end{subfigure}
    \begin{subfigure}[t]{0.35\textwidth}
        \includegraphics[width=\textwidth]{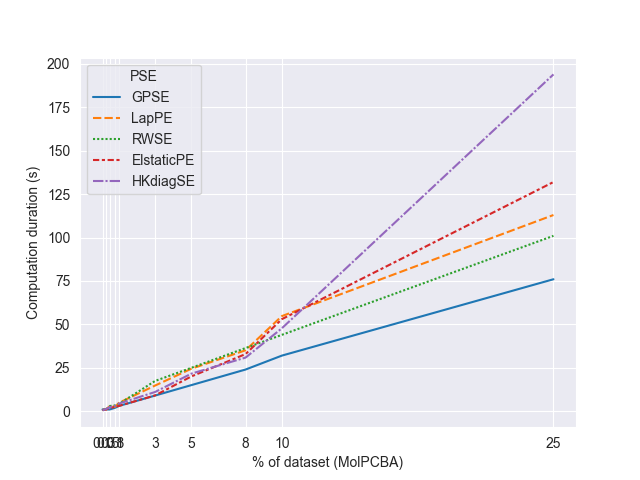}
        \caption{\method + individual PSEs only}
    \end{subfigure}
    \begin{subfigure}[t]{0.27\textwidth}
        \includegraphics[width=\textwidth]{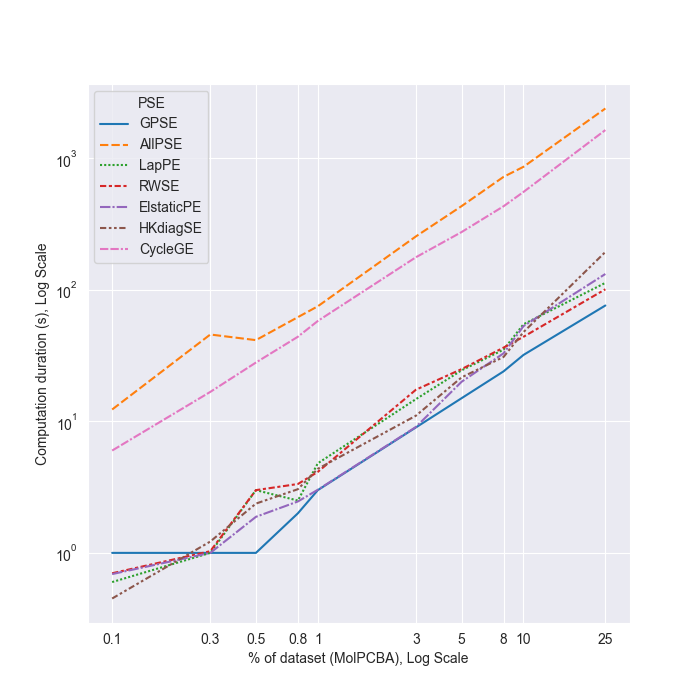}
        \caption{Log-log plot of \method + individual PSEs + combined PSEs (AllPSE)}
    \end{subfigure}
    \vspace{-8pt}
    \caption{Scaling of PSE computation time with respect to number of graphs as \% of MolPCBA dataset used. Visualization of Table~\ref{tab:graph_number_scaling}.}
    \label{fig:graph_number_scaling}
\end{figure}

\input{tables/graph_size_scaling}

\begin{figure}[!htp]
    \centering
    \begin{subfigure}[t]{0.56\textwidth}
        \includegraphics[width=\textwidth]{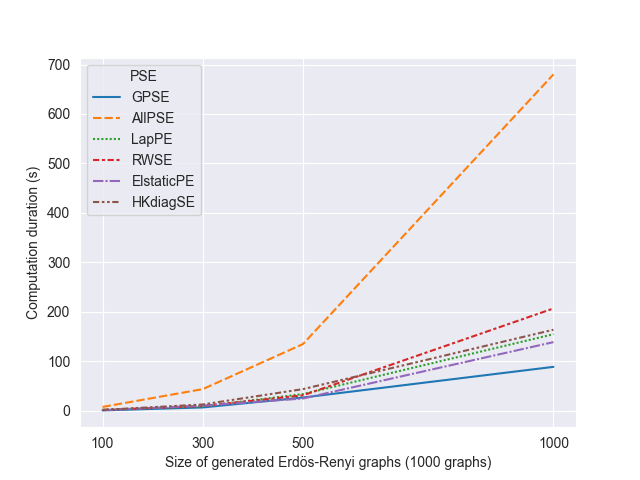}
        \caption{GPSE + individual PSEs + combined PSEs (AllPSE)}
    \end{subfigure}
    \begin{subfigure}[t]{0.42\textwidth}
        \includegraphics[width=\textwidth]{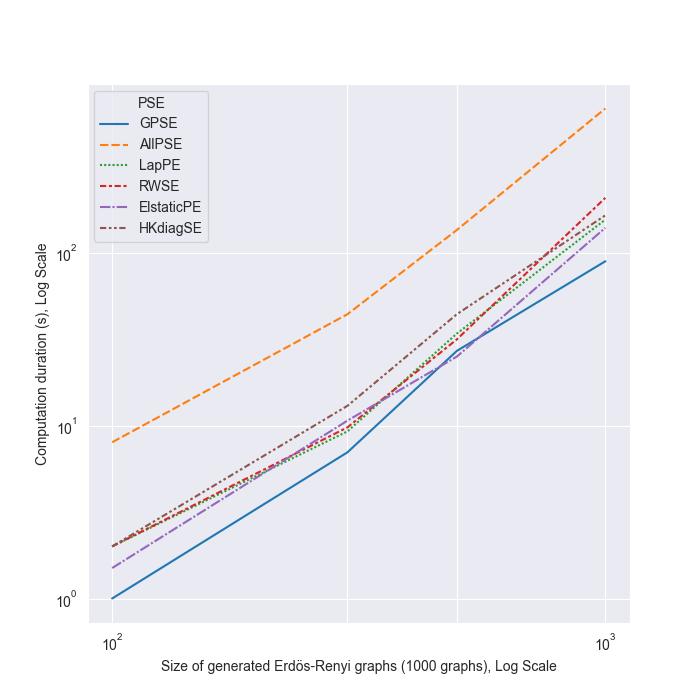}
        \caption{Log-log plot of GPSE + individual PSEs + combined PSEs (AllPSE)}
    \end{subfigure}
    \vspace{-5pt}
    \caption{Scaling experiments with respect to size of graphs, keeping the number of graphs in each dataset constant. Visualization of Table~\ref{tab:graph_size_scaling}.}
    \label{fig:graph_size_scaling}
\end{figure}

In these experiments, we measure and compare the computation time of GPSE with those of individual PSEs used in the pre-training of the GPSE model, as well as their combination (AllPSE). We conducted two sets of experiments. In the first, we used a dataset of similarly sized graphs in MolPCBA, but ran PSE computation for an increasing percentage of the dataset (Figure~\ref{fig:graph_number_scaling}). In the second, we generated multiple datasets of 1000 synthetic (Erd\H{o}s-Rényi) graphs, scaling up the number of nodes per graph (Figure~\ref{fig:graph_size_scaling}) in each.

In both sets of experiments, GPSE is considerably faster to compute than the individual PSEs, and orders-of-magnitude faster than computing and combining all PSEs as AllPSE. Additionally, we observed that GPSE scales better than individual and combined PSEs. The better scaling properties of GPSE are particularly evident in scaling graph sizes (Figure~\ref{fig:graph_size_scaling}): As we scaled up the number of nodes in a graph to 1000, the advantage of GPSE became more apparent. This is somewhat an expected result: Regardless of graph size, inference of GPSE involves $O(Lm)$ computations, where L is the number of layers, and m is the number of edges; thus GPSE scales linearly on the number of edges in a graph. On the other hand, LapPE, for example, is expected to have polynomial complexity, as it involves eigendecomposing the graph Laplacian.

Another important point to highlight is that at inference time, \emph{the complexity of GPSE remains unchanged} regardless of how many types of PSEs were used to train the model. This leads to significant advantages over AllPSE, which relies on computing and concatenating all PSEs. The scalability issues of AllPSE is additionally exacerbated when useful but highly expensive PSEs such as CycleGE are used.
\color{black}

\newpage
\section{Additional results}

% \begin{figure}[!htp]
%     \centering
%     \begin{minipage}{0.45\textwidth}
%         \centering
%         \includegraphics[width=0.9\textwidth]{figures/abl_conv_layers_vn.pdf}
%         \caption{Virtual node (VN), convolution type, and layers ablation using 5\% MolPCBA for training \method.}
%   \label{fig:abl_conv_layers_vn}
%     \end{minipage}\hfill
%     \begin{minipage}{0.5\textwidth}
%         \centering
%         \includegraphics[width=0.9\textwidth]{figures/scaling_law.pdf}
%         \caption{Training size scaling law for \method on MolPCBA.}
%   \label{fig:scaling_law}
%     \end{minipage}
% \end{figure}

\begin{figure}[!htpb]
    \centering
    \includegraphics[width=0.6\textwidth]{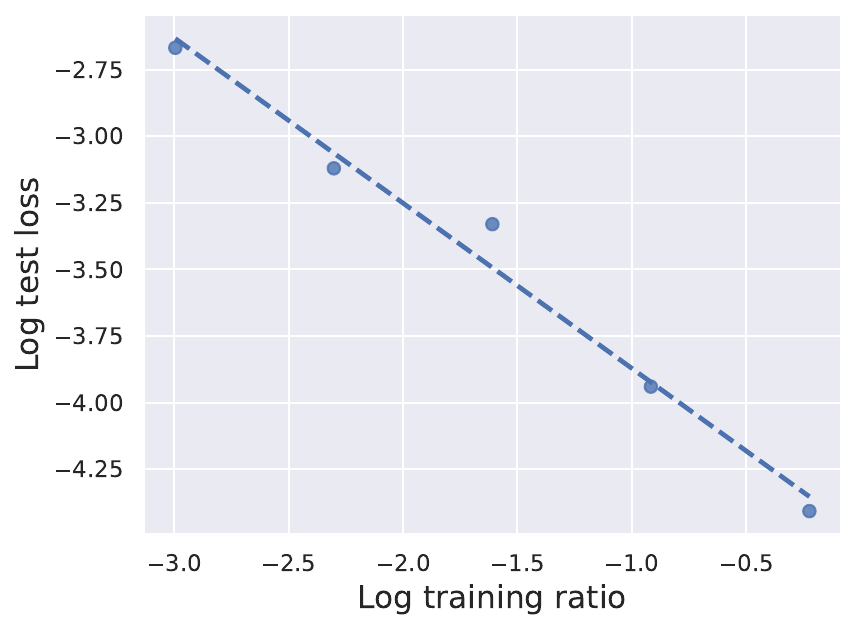}
    \caption{Training size scaling law for \method on MolPCBA.}
    \label{fig:scaling_law}
\end{figure}

\input{tables/abl_pse_perf}
\input{tables/abl_pcqm4m_pse}

\input{tables/abl_pt}

% \input{tables/abl_conv_layers_vn}

\input{tables/abl_pt_task}

\input{tables/abl_pt_dataset}

\input{tables/abl_gpse_lap}

\input{tables/molnet_full}

\newpage
\section{Additional discussions}
\label{appendix:discussion}

\textbf{Why does \method improve over precomputed PSEs? } Our results demonstrate that \method encodings can improve upon augmenting GNNs with precomputed PSEs in downstream tasks. The fact that we can \emph{recover} the target PSEs in pretraining (Table~\ref{tab:pse_perf}) accounts for why we can match the original PSEs. Why we \emph{improve} upon them, meanwhile, can be attributed to our joint encoding:
% Previous work only considers one encoding at a type (e.g. LapPE or RWSE) while our joint representation is able to incorporate multiple families of PSEs (\S\ref{sec:target_pse}). \rl{I would not explicitly mention the improvement is because we are using multiple PSE like it here. It begs the question: why not just use multiple PSE to begin with?} Additionally, our high-dimensional joint representation is richer than simply predicting the target PSEs, and likely encodes relationships between PSEs as well.
Learning to encode a diverse collection of PSEs leads to a general embedding space that abstracts both local and global perspectives of the query graph, which are more readily useable by the downstream model compared to the unprocessed PSEs.
% TODO: add discussion on why GPSE is better than AllPSE here
This also explains why a joint encoding outperforms the concatenation of all encodings, AllPSE. Concatenating many encodings very likely to leads to redundant representations, and also introduces significant noise to node features particularly in datasets/downstream tasks where only a small portion of the encodings are useful: This is reflected in most experiments where not only GPSE, but also RWSE- or LapPE-only configurations outperform AllPSE.
%Why \method can \emph{improve} on them can be attributed to a number of things. Primarily, we are able to learn a joint encoding for multiple PSEs, as opposed to prior work which only considers one encoding at a type (e.g. LapPE or RWSE), and selects the best one for each dataset. While it is true that not all encodings are equally useful on a given graph, learning a joint representation means we are able to extract information from \emph{all} PSEs considered including the ``best'' one. Our representations are richer and likely encode relationships between PSEs as well as the information to predict them, thus explaining the superior performance of \method and the SOTA results.

\textbf{What advantages does \method bring over traditional SSL pre-training? }
% \rl{possibly suggest how \method can be used together with traditional SSL to even further improve performance}
% Our results show that \method trained on MolPCBA leads to substantial downstream performance improvements across small MoleculeNet datasets (in addition to achieving best performance on ToxCast and SIDER), while most SSL pre-training methods suffer from substantial negative transfer across the datasets. These results are further amplified when we consider the comparative disadvantage of \method: Our encoder extracts only positional and structural information, and does not have any access to node features which other methods can leverage. On why \method is more stable across datasets than most SSL methods, we turn to the SSL training objectives: Contrastive SSL approaches we test against impose certain invariances (e.g. to certain graph structure or node feature perturbations) by design, which leads to limited improvements or negative transfer when these invariances do not hold downstream. \method encodings we learn are instead intrinsic to the graphs, and thus transfer more reliably as they do not rely on such extrinsic invariances.
In addition to being less prone to negative transfer and having competitive performance (Table~\ref{tab:molnet}), \method provides a few more advantages over traditional SSL pre-training methods: (1) \method uses randomly generated features instead of dataset-specific graph features, thus can be applied to arbitrary graph datasets; (2) \method is only used as a feature extractor and hence does not impose any constraint on the downstream model. Despite these differences, \method can be complementary to traditional SSL to further enhance the prediction performance, for example, by using \method encodings as input features to the SSL pre-training.

\textbf{Why is \method transferable to OOD data? }
The transferability of \method to OOD data is uncommon in the graph SSL pre-training literature, particularly for molecular applications.
% As discussed, SSL encodings are useful only when the pretraining and downstream dataset and tasks are well-aligned, limiting its application domains. This is exemplified by most graph SSL papers using only MoleculeNet datasets for benchmarking; the models are usually pre-trained on the larger MolPCBA or PPI datasets, and tested downstream on the smaller ones as in Table~\ref{tab:molnet}. \method instead shows remarkably superior transferability
We hypothesize that \method's transferability is a consequence of the choice of its predictive self-supervision tasks, which contain a mixture of both local and global intrinsic graph information. This encourages \method to capture global invariances using local information, hence allowing it to extract valuable representations on graphs that are different in sizes and connectivity from the training graphs.
% This result also complements \citet{Hu*2020Strategies}, which showed that pre-training at both node and graph levels leads to better SSL transferability.

% - subgraph structures (local structural encodings); what about global ones though..

\textbf{When does \method help, and when does it not? }
\method provides essential information to the model when the downstream task requires positional or structural information of the graph or better node identifiability in general, which is typically the case for molecular property predictions~\citep{Hu*2020Strategies}. Conversely, for downstream tasks that do not rely on such information, e.g. protein function prediction using the protein interaction network (Table~\ref{tab:nodebench_full}), the benefits from \method are not as apparent.

%% file: tables/hp_mol_bench.tex
\begin{table}[!htp]\centering
\caption{GPS+\method hyperparameters for molecular property prediction benchmarks}\label{tab:hp_mol_bench}
\footnotesize
\begin{tabular}{lcccc}\toprule
Hyperparameter &\textbf{ZINC} (subset) &\textbf{PCQM4Mv2} (subset) &\textbf{MolHIV} &\textbf{MolPCBA} \\
\midrule
\# GPS Layers &10 &5 &10 &5 \\
Hidden dim &64 &304 &64 &384 \\
GPS-MPNN &GINE &GatedGCN &GatedGCN &GatedGCN \\
GPS-SelfAttn &-- &Transformer &Transformer &Transformer \\
\# Heads &4 &4 &4 &4 \\
Dropout &0.00 &0.00 &0.05 &0.20 \\
Attention dropout &0.50 &0.50 &0.50 &0.50 \\
Graph pooling &mean &mean &mean &mean \\
\midrule
PE dim &32 &128 &20 &48 \\
PE encoder &2-Layer MLP &2-Layer MLP &Linear &Linear \\
Input dropout &0.50 &0.50 &0.30 &0.30 \\
Output dropout &0.00 &0.20 &0.10 &0.10 \\
Batchnorm &yes &no &yes &yes \\
\midrule
Batch size &32 &256 &32 &512 \\
Learning rate &0.001 &0.0002 &0.0001 &0.0005 \\
\# Epochs &2000 &100 &100 &100 \\
\# Warmup epochs &50 &5 &5 &5 \\
Weight decay &1.00e-5 &1.00e-6 &1.00e-5 &1.00e-5 \\
\midrule
\# Parameters &292,513 &6,297,345 &573,025 &9,765,264 \\
PE precompute &2 min &1.5 hr &8 min &1.3 hr \\
Time (epoch/total) &10s/5.78h &102s/2.82h &121s/3.37h &185s/5.15h \\
\bottomrule
\end{tabular}
\end{table}

%% file: tables/hp_transfer_bench.tex
\begin{table}[!htp]\centering
\caption{GPS+\method hyperparameters for transferability benchmarks}\label{tab:hp_transfer_bench}
\footnotesize
\begin{tabular}{lcccc}\toprule
Hyperparameter &\textbf{Peptides-struct} &\textbf{Peptides-func} &\textbf{CIFAR10} &\textbf{MNIST} \\
\midrule
\# GPS Layers &4 &4 &3 &3 \\
Hidden dim &96 &96 &52 &52 \\
GPS-MPNN &GatedGCN &GatedGCN &GatedGCN &GatedGCN \\
GPS-SelfAttn &Transformer &Transformer &Transformer &Transformer \\
\# Heads &4 &4 &4 &4 \\
Dropout &0.00 &0.00 &0.00 &0.00 \\
Attention dropout &0.50 &0.50 &0.50 &0.50 \\
Graph pooling &mean &mean &mean &mean \\
\midrule
PE dim &8 &24 &8 &8 \\
PE encoder &Linear &2-Layer MLP &2-Layer MLP &Linear \\
Input dropout &0.10 &0.10 &0.30 &0.50 \\
Output dropout &0.05 &0.00 &0.00 &0.00 \\
Batchnorm &yes &yes &no &no \\
\midrule
Batch size &128 &128 &16 &16 \\
Learning rate &0.0005 &0.0003 &0.001 &0.001 \\
\# Epochs &200 &200 &100 &100 \\
\# Warmup epochs &10 &10 &5 &5 \\
Weight decay &1.00e-4 &0 &1.00e-5 &1.00e-4 \\
\midrule
\# Parameters &510,435 &529,250 &120,886 &119,314 \\
PE precompute &3 min &3 min &14 min &16 min \\
Time (epoch/total) &12s/0.65h &12s/0.67h &88s/2.44h &104s/2.90h \\
\bottomrule
\end{tabular}
\end{table}

%% file: tables/hp_node_bench.tex
\begin{table}[!htp]\centering
\caption{Downstream MPNN hyperparameters for node-level benchmarks.}\label{tab:hp_node_bench}
\footnotesize
\begin{tabular}{lcc}\toprule
Hyperparameter &\textbf{arXiv} &\textbf{Proteins} \\
\midrule
MPNN &SAGE &GATEv2 \\
\# MPNN Layers &3 &3 \\
Hidden dim &256 &256 \\
Dropout &0.50 &0.00 \\
\midrule
PE dim &32 &32 \\
PE encoder &2-Layer MLP &Linear \\
Input dropout &0.50 &0.40 \\
Output dropout &0.20 &0.05 \\
Batchnorm &no &no \\
\midrule
% Sampling &Full &Neighbors \\
Learning rate &0.01 &0.01 \\
\# Epochs &500 &1000 \\
Weight decay &0 &0 \\
\midrule
\# Parameters &534,888 &910,448 \\
PE precompute &8 sec &3 min \\
Time (epoch/total) &0.25s/0.07h &35s/9.40h \\
\bottomrule
\end{tabular}
\end{table}

%% file: tables/hp_molnet_and_wl.tex
\begin{table}[ht]
    \caption{\method processing encoder hyperparameters for MoleculeNet small benchmarks and synthetic WL graph benchmarks.}
    \label{tab:training_params}
    \begin{subtable}[h]{0.45\textwidth}\centering
    \caption{MoleculeNet small benchmarks settings.}\label{tab:hp_molnet_small_bench}
    \begin{tabular}{lc}\toprule
    Hyperparameter & \\
    \midrule
    PE dim &64 \\
    PE encoder &Linear \\
    Input dropout &0.30 \\
    Output dropout &0.10 \\
    Batchnorm &yes \\
    \midrule
    Learning rate &0.003 \\
    \# Epochs &100 \\
    \# Warmup epochs &5 \\
    Weight decay &0 \\
    \bottomrule
    \end{tabular}
    \end{subtable}
    \hfill
    \begin{subtable}[h]{0.45\textwidth}\centering
    \caption{Synthetic WL graph benchmarks settings.}\label{tab:hp_wl_bench}
    \begin{tabular}{lc}\toprule
    Hyperparameter & \\
    \midrule
    PE dim &128 \\
    PE encoder &Linear \\
    Input dropout &0.00 \\
    Output dropout &0.00 \\
    Batchnorm &yes \\
    \midrule
    Learning rate &0.002 \\
    \# Epochs &200 \\
    Weight decay &0 \\
    \bottomrule
    \end{tabular}
    \end{subtable}
\end{table}

%% file: tables/dataset_info.tex
\begin{table}[!htp]\centering
\caption{Task information for datasets used in transferability experiments.}
\label{table:dataset_info}
\footnotesize
\begin{tabular}{@{}lrrrccrc@{}}
\toprule
\multirow{2}{*}{Dataset} & \multicolumn{1}{c}{Num.}   & \multicolumn{1}{c}{Num.}   & \multicolumn{1}{c}{Num.}   & \multicolumn{1}{c}{Pred.} & \multicolumn{1}{c}{Pred.}                   & \multicolumn{1}{c}{Num.}  & \multirow{2}{*}{Metric} \\
                         & \multicolumn{1}{c}{graphs} & \multicolumn{1}{c}{nodes}  & \multicolumn{1}{c}{edges}  & \multicolumn{1}{c}{level} & \multicolumn{1}{c}{task}                    & \multicolumn{1}{c}{tasks} &                         \\ \midrule
ZINC-subset              & 12,000                     & 23.15  & 24.92  & graph                     & reg.              & 1     & MAE                     \\
CIFAR10                  & 60,000                     & 117.63 & 469.10 & graph                     & class. (10-way) & 1     & ACC                \\
MNIST                    & 70,000                     & 70.57  & 281.65 & graph                     & class. (10-way) & 1     & ACC                \\
MolHIV                   & 41,127                     & 25.51  & 27.46  & graph                     & class. (binary) & 1     & AUROC                 \\
MolPCBA                  & 437,929                    & 25.97  & 28.11  & graph                     & class. (binary) & 128   & AP                      \\
MolBBBP                  & 2,039                      & 24.06  & 25.95  & graph                     & class. (binary) & 1     & AUROC                 \\
MolBACE                  & 1,513                      & 34.09  & 36.86  & graph                     & class. (binary) & 1     & AUROC                 \\
MolTox21                 & 7,831                      & 18.57  & 19.29  & graph                     & class. (binary) & 21    & AUROC                 \\
MolToxCast               & 8,576                      & 18.78  & 19.26  & graph                     & class. (binary) & 617   & AUROC                 \\
MolSIDER                 & 2,039                      & 33.64  & 35.36  & graph                     & class. (binary) & 27    & AUROC                 \\
PCQM4Mv2-subset          & 446,405                    & 14.15  & 14.58  & graph                     & reg.              & 1     & MAE                     \\
Peptides-func            & 15,535                     & 150.94 & 153.65 & graph                     & class. (binary) & 10    & AP                      \\
Peptides-struct          & 15,535                     & 150.94 & 153.65 & graph                     & reg.              & 11    & MAE                     \\
CSL                      & 150                        & 41.00  & 82.00  & graph                     & class. (10-way) & 1     & ACC                \\
EXP                      & 1,200                      & 48.70  & 60.44  & graph                     & class. (binary) & 1     & ACC                \\
arXiv                    & 1                          & 169K   & 40M    & node                      & class. (40-way) & 1     & ACC                \\
Proteins                 & 1                          & 133K   & 1.2M   & node                      & class. (binary) & 112   & AUROC                 \\ \bottomrule
\end{tabular}
\end{table}

%% file: tables/dataset_stats.tex
\begin{table}[!htp]\centering
\caption{Classical graph properties of graph-level datasets used in transferability experiments.}
\label{table:dataset_props}
% \fontsize{8pt}{8pt}\selectfont 
\setlength\tabcolsep{4pt} % default value: 6pt
\footnotesize
\begin{adjustwidth}{-2.5 cm}{-2.5 cm}\centering
%\rowcolors{3}{}{lightgray}
\renewcommand{\arraystretch}{1.2}
\begin{tabular}{lrrrrrrrrrr}\toprule
    &Num. &Num. &\multirow{2}{*}{Density} &\multirow{2}{*}{Connectivity} &\multirow{2}{*}{Diameter} &Approx. &\multirow{2}{*}{Centrality} &Cluster. &Num. \\
&nodes &edges & & & &max clique & &coeff. &triangles \\\midrule
    ZINC-subset &23.15 &24.92 &0.101 &1.00 &12.47 &2.06 &0.101 &0.006 &0.06\\
    CIFAR10 &117.63 &469.10 &0.068 &3.56 &9.14 &5.65 &0.068 &0.454 &502.66\\
    MNIST &70.57 &281.65 &0.116 &3.71 &6.83 &5.56 &0.116 &0.478 &316.65\\
    MolHIV &25.51 &27.46 &0.103 &0.927 &11.06 &2.02 &0.103 &0.002 &0.03\\
    MolPCBA &25.97 &28.11 &0.093 &0.998 &13.56 &2.02 &0.093 &0.002 &0.02\\
    MolBBBP &24.06 &25.95 &0.114 &0.950 &10.75 &2.03 &0.114 &0.003 &0.03\\
    MolBACE &34.09 &36.86 &0.070 &1.00 &15.22 &2.10 &0.070 &0.007 &0.10\\
    MolTox21 &18.57 &19.29 &0.157 &0.976 &9.37 &2.02 &0.159 &0.003 &0.03\\
    MolToxCast &18.78 &19.26 &0.154 &0.803 &7.57 &2.02 &0.154 &0.003 &0.03\\
    MolSIDER &33.64 &35.36 &0.103 &0.856 &12.45 &2.02 &0.120 &0.004 &0.04\\
    PCQM4Mv2-subset &14.15 &14.58 &0.163 &1.00 &7.95 &2.06 &0.163 &0.010 &0.07\\
    Peptides-func &150.94 &153.65 &0.022 &0.990 &56.42 &2.00 &0.022 &0.000 &0.001\\
    Peptides-struct &150.94 &153.65 &0.022 &0.990 &56.42 &2.00 &0.022 &0.000 &0.001\\
    CSL &41.00 &82.00 &0.100 &3.98 &6.00 &2.10 &0.100 &0.050 &4.10\\
    EXP &48.70 &60.44 &0.054 &0.00 &1.00 &2.00 &0.054 &0.000 &0.00\\
    %Arxiv &169K &40M & & & & & & &\\
    %Proteins &133K &1.2M & & & & & & &\\
\bottomrule
\end{tabular}
\end{adjustwidth}
\end{table}

%% file: tables/gpse_train_time.tex
\begin{table}[!htp]\centering
\caption{GPSE training times. Target PSE pre-computation included.}
\label{tab:gpse_train_time}
\vspace{3pt}
% \scriptsize
\begin{tabular}{@{}lcccc@{}}\toprule
\textbf{Training dataset} & \textbf{Num. unique graphs} & \textbf{Target PSE pre-comp time} & \textbf{Time (epoch/total)} & \textbf{Full training time} \\ \midrule
MolPCBA (default)   & 323,555   & 1.58h &  596s / 19.88h & 21.46h \\
PCQM4Mv2-full       & 273,920   & 0.87h &  429s / 14.30h & 15.17h \\
ZINC-full           & 219,384   & 0.89h &  398s / 13.26h & 14.15h \\
GEOM                & 169,925   & 0.78h &  321s / 10.69h & 11.47h \\
ChEMBL              & 970,963   & 5.97h & 2509s / 83.65h & 89.62h \\
\bottomrule
\end{tabular}
\end{table}

%% file: tables/gpse_test_time.tex
\begin{table}[!htp]\centering
% \captionsetup{width=.8\linewidth}
\caption{GPSE inference times. LapPE and RWSE computation times are included for comparison. Missing entries are due to experimental settings not included in the benchmarking experiments.}
\label{tab:gpse_test_time}
\vspace{3pt}
% \scriptsize
\begin{tabular}{@{}lrrrr@{}}\toprule
\textbf{Dataset} & \textbf{Num. graphs} & \textbf{Time (\method)} & \textbf{Time (LapPE)} & \textbf{Time (RWSE)} \\ \midrule
ZINC-subset      & 12,000               & 6 sec                & 25 sec                & 11 sec               \\
PCQM4Mv2-subset  & 446,405              & 3.57 min             & 3.88 min              & 7.32 min             \\
PCQM4Mv2-full    & 3,746,620            & 31.15 min            & --                    & 51 min *             \\
MolHIV           & 41,127               & 23 sec               & 37 sec                & 58 sec *             \\
MolPCBA          & 437,929              & 4.6 min              & 6.13 min              & 8.33 min *           \\
Peptides         & 15,535               & 28 sec               & 73 sec *              & --                   \\
CIFAR10          & 60,000               & 2.15 min             & 2.55 min *            & --                   \\
MNIST            & 70,000               & 100 sec              & 96 sec *              & --                   \\
arXiv            & 1                    & 4 sec                & --                    & --                   \\
Proteins         & 1                    & 6.68 min **          & --                    & --                   \\ \bottomrule
\multicolumn{5}{l}{* Obtained from the GPS paper}\\
\multicolumn{5}{l}{** Neighbor batched computation (batch size: 1024, neighbor sizes: 30, 20, 10, 5, …, 5)}\\
\end{tabular}
\end{table}

%% file: tables/graph_number_scaling.tex
\begin{table}[!ht]
    \centering
    % \scriptsize
    \caption{Runtimes of each PSE computation with respect to percentage of dataset used.}\label{tab:graph_number_scaling}
    \footnotesize
    \begin{tabular}{lrrrrrrrrrrr}
    \toprule
        \textbf{PSE / \% MolPCBA} & \textbf{0.1\%} & \textbf{0.3\%} & \textbf{0.5\%} & \textbf{0.8\%} & \textbf{1\%} & \textbf{3\%} & \textbf{5\%} & \textbf{8\%} & \textbf{10\%} & \textbf{25\%} \\ \midrule
        \textbf{\method} & 1s & 1s & 1s & 2s & 3s & 9s & 15s & 24s & 32s & 1m 16s \\
        AllPSE & 12s & 46s & 41s & 1m 2s & 1m 15s & 4m 15s & 7m 13s & 12m 3s & 14m 22s & 39m 52s \\
        LapPE & 1s & 1s & 3s & 3s & 5s & 15s & 24s & 35s & 55s & 1m 53s \\
        RWSE & 1s & 1s & 3s & 3s & 4s & 17s & 25s & 36s & 44s & 1m 41s \\
        ElstaticPE & 1s & 1s & 2s & 2s & 3s & 10s & 20s & 33s & 53s & 2m 12s \\
        HKdiagSE & 1s & 1s & 2s & 3s & 4s & 11s & 22s & 31s & 48s & 3m 14s \\
        CycleGE & 6s & 17s & 28s & 44s & 58s & 2m 57s & 4m 35s & 7m 12s & 9m 15s & 27m 20s \\
        \bottomrule
    \end{tabular}
\end{table}

%% file: tables/graph_size_scaling.tex
\begin{table}[!ht]
    \centering
    % \scriptsize
    \caption{Runtimes of each PSE computation with respect to average graph sizes in dataset. CycleGE is excluded both in itself and as part of AllPSE, as cycle counting on large, dense and regular Erd\H{o}s-Rényi graphs become computationally infeasible.}\label{tab:graph_size_scaling}
    \footnotesize
    \begin{tabular}{lrrrr}
    \toprule
        \textbf{PSE / Graph size} & \textbf{100} & \textbf{300} & \textbf{500} & \textbf{1000} \\ \midrule
        \textbf{\method} & 1s & 7s & 27s & 1m 29s \\
        AllPSE (No CycleGE) & 8s & 44s & 2m 15s & 11m 20s \\
        LapPE & 2s & 9s 250ms & 34s & 2m 35s \\
        RWSE & 2s & 9s 760ms & 31s 480ms & 3m 27s \\
        ElstaticPE & 1s 500ms & 10s 670ms & 25s 30ms & 2m 19s \\
        HKdiagSE & 2s & 13s & 44s & 2m 44s \\
        \bottomrule
    \end{tabular}
\end{table}

%% file: tables/abl_pse_perf.tex
\begin{table}[!htp]\centering
\caption{\method architecture ablation. Held-out positional and structural encodings prediction performance when trained on 5\% MolPCBA (\Rsq scores $\uparrow$). Ablated settings of the \method architecture are listed and compared to the full \method settings. The performance of the full \method architecture is shown in the bottom row.
% \lr{Here it is not clear what the GPSE model looks like/how it compares to the ablated arch. choices.}\rl{Does adding the ablated option column vs. GPSE option column help?}
}
\label{tab:abl_pse_perf}
\vspace{3pt}
% \scriptsize
\footnotesize
%% Make the tables more readable if we have the space.
% \fontsize{8.5pt}{8.5pt}\selectfont
% \renewcommand{\arraystretch}{1.1}
% \setlength\tabcolsep{5pt} % default value: 6pt
% \tiny
\hspace*{-0.5cm}
\begin{tabular}{ll|ccccccc}\toprule
Ablated setting &\method default & Overall &ElstaticPE &LapPE &RWSE &HKdiagSE &EigValSE &CycleSE \\
\midrule[0.75pt]
10 layers &20 layers & 0.9585 & 0.9376 & 0.9302 & 0.9645 & 0.9622 & 0.9543 & 0.9701 \\
128 dim & 512 dim & 0.9688 & 0.9484 & 0.9501 & 0.9729 & 0.9734 & 0.9706 & 0.9739 \\
\midrule
GCN & GatedGCN &0.0409 &0.0408 &0.0325 &0.0424 &0.0396 &0.0483 &0.0410\\
GIN & GatedGCN & 0.6095 & 0.6953 & 0.4180 & 0.6237 & 0.6349 & 0.4002 & 0.6391 \\
GATv2 & GatedGCN &0.9580 &0.9560 &0.9476 &0.9643 &0.9530 &0.9679 &0.9561 \\
\midrule
No VN &VN & 0.9478 & 0.9340 & 0.9359 & 0.9552 & 0.9479 & 0.9314 & 0.9568 \\
Shared MLP head &Indep. MLP heads & 0.9751 & 0.9619 & 0.9644 & 0.9802 & 0.9764 & 0.9714 & \textbf{0.9778} \\
% &Constant feat. &0.9779 &0.xxxx &0.xxxx &0.xxxx &0.xxxx &0.xxxx &0.xxxx \\
\midrule
\multicolumn{2}{c|}{\method} &  \textbf{0.9790} &    \textbf{0.9638} &  \textbf{0.9725} &  \textbf{0.9837} &  \textbf{0.9808} &  \textbf{0.9818} &  0.9774 \\
% &\method &\textbf{0.9979} &\textbf{0.xxxx} &\textbf{0.xxxx} &\textbf{0.xxxx} &\textbf{0.xxxx} &\textbf{0.xxxx} &\textbf{0.xxxx} \\
\bottomrule
\end{tabular}
\end{table}

%% file: tables/abl_pcqm4m_pse.tex
\begin{table}[!htp]\centering
\caption{Five PSE augmentations combined with five different GNN models evaluated on the PCQM4Mv2-subset dataset. Performance is evaluated as mean absolute error (MAE $\downarrow$) and averaged over 4 seeds. \textcolor{dark2pink}{Red} indicates worse than baseline (none) performance.}\label{tab:abl_pcqm4m_pse}
\vspace{3pt}
% \scriptsize
\footnotesize
\begin{tabular}{lcccccc}
\toprule
&GCN &GatedGCN &GIN &GINE &Transformer &Avg. reduction \\
\midrule
none &0.1934 ± 0.0012 &0.1845 ± 0.0031 &0.1790 ± 0.0011 &0.1364 ± 0.0011 &0.4193 ± 0.0167 & -- \\
\midrule
rand &\textcolor{dark2pink}{0.7604 ± 0.0019} &\textcolor{dark2pink}{0.7515 ± 0.0027} &\textcolor{dark2pink}{0.7532 ± 0.0021} &\textcolor{dark2pink}{0.4269 ± 0.0068} &\textcolor{dark2pink}{0.9810 ± 0.0064} &N/A \\
LapPE &0.1834 ± 0.0023 &0.1757 ± 0.0010 &0.1720 ± 0.0018 &0.1338 ± 0.0006 &0.2433 ± 0.0056 &\cellcolor[HTML]{fff2cc}11.55\% \\
RWSE &0.1877 ± 0.0025 &0.1782 ± 0.0012 &0.1695 ± 0.0007 &0.1317 ± 0.0005 &0.1930 ± 0.0016 &\cellcolor[HTML]{ffe6a1}13.82\% \\
LapPE+RWSE	&0.1895 ± 0.0022 &0.1632 ± 0.0016 &0.1705 ± 0.0016 &\textcolor{dark2pink}{0.1370 ± 0.0005} &0.1884 ± 0.0011 &\cellcolor[HTML]{ffe498}14.76\%\\
AllPSE	&0.1903 ± 0.0010 &0.1595 ± 0.0014 &0.1656 ± 0.0004 &\textcolor{dark2pink}{0.1416 ± 0.0040} &0.2133 ± 0.062 &\cellcolor[HTML]{ffe8a8}13.59\%\\
\textbf{\method} &\textbf{0.1822 ± 0.0028} &\textbf{0.1495 ± 0.0015} &\textbf{0.1615 ± 0.0015} &\textbf{0.1294 ± 0.0006} &\textbf{0.1805 ± 0.0021} &\cellcolor[HTML]{ffd666}\textbf{19.32\%} \\
\bottomrule
\end{tabular}
\end{table}

%% file: tables/abl_pt.tex
\begin{table}[htp]
    \caption{\method pre-training ablations.}
    \begin{subtable}[h]{0.45\textwidth}\centering
    % \scriptsize
    % \fontsize{8.5pt}{8.5pt}\selectfont
    \footnotesize
    \renewcommand{\arraystretch}{1.2}
    \setlength\tabcolsep{6pt} % default value: 6pt
    \caption{Virtual node, convolution type, and layers ablation using 5\% MolPCBA for training.}\label{tab:abl_conv_layers_vn}
    % \begin{tabular}{@{}ccccc@{}}
    \begin{tabular}{@{\extracolsep{10pt}}ccccc@{}}
    \toprule
    &\multicolumn{2}{c}{\method (GatedGCN)} &\multicolumn{2}{c}{\method (GIN)} \\
    \cmidrule{2-3}
    \cmidrule{4-5}
    Layers &VN & no VN &VN & no VN \\
    \midrule
    5 &0.8387 &0.6982 &0.4879 &0.1347 \\
    10 &0.9585 &0.8353 &0.5156 &0.2476 \\
    15 &0.9716 &0.9231 &0.5887 &0.2523 \\
    20 &0.9778 &0.9478 &0.6095 &0.2743 \\
    30 &0.9806 &0.9559 &0.4149 &0.3740 \\
    40 &0.9782 &0.9459 &0.5420 &0.3968 \\
    \bottomrule
    \end{tabular}
    \end{subtable}
    \hfill
    \begin{subtable}[h]{0.45\textwidth}\centering
    % \scriptsize
    % \fontsize{8.5pt}{8.5pt}\selectfont
    \footnotesize
    \renewcommand{\arraystretch}{1.2}
    \setlength\tabcolsep{6pt} % default value: 6pt
    \caption{Training size scaling law for \method on MolPCBA.}\label{tab:scaling_law}
    \begin{tabular}{@{}cc@{}}
    \toprule
    Training size & Overall test loss (MAE + cosine loss) $\downarrow$\\
    \midrule
    5\%     & 0.06939 \\
    10\%    & 0.04414 \\
    20\%    & 0.03579 \\
    40\%    & 0.01945 \\
    80\%    & 0.01219 \\
    \bottomrule
    \end{tabular}
    \end{subtable}
\end{table}

%% file: tables/abl_pt_task.tex
% \begin{table}[!htp]\centering
% \scriptsize
% \caption{GPSE training task ablation}\label{tab:abl_pt_task}
% \begin{tabular}{@{}cc@{}}
% \toprule
% Excluded task & \textbf{PCQM4Mv2} (subset)\\
% \midrule
% --                  & \textbf{0.1196 ± 0.0004} \\
% \midrule
% LapPE \& EigVals    & 0.1200 ± 0.0006 \\
% ElstaticPE          & \underline{0.1197 ± 0.0007} \\
% RWSE                & 0.1205 ± 0.0006 \\
% HKdiagSE            & 0.1202 ± 0.0004 \\
% CycleSE             & \underline{0.1199 ± 0.0011} \\
% \bottomrule
% \end{tabular}
% \end{table}

% \begin{wraptable}{r}{0.48\linewidth}\centering
% \vspace{-0.25in}
\begin{table}\centering
% \scriptsize
% \fontsize{8.5pt}{8.5pt}\selectfont
\renewcommand{\arraystretch}{1.2}
\setlength\tabcolsep{6pt} % default value: 6pt
\caption{\method training task ablation. The colors indicate whether a particular PSE task for training \method \textcolor{dark2green}{improves} or \textcolor{dark2pink}{worsens} the downstream performance.}\label{tab:abl_pt_task}
\footnotesize
\begin{tabular}{@{}ccc@{}}
\toprule
& \textbf{PCQM4Mv2} (subset) & \textbf{ogbg-molhiv} \\
Excluded task &\textbf{MAE} $\downarrow$ &\textbf{AUROC} $\uparrow$ \\
\midrule
--                  & 0.1196 ± 0.0004 & 0.7815 ± 0.0133 \\
\midrule
LapPE \& EigVals    & \textcolor{dark2green}{0.1200 ± 0.0006}    & \textcolor{dark2pink}{0.7849 ± 0.0067} \\
ElstaticPE          & \textcolor{black}{0.1197 ± 0.0007}        & \textcolor{dark2green}{0.7681 ± 0.0146} \\
RWSE                & \textcolor{dark2green}{0.1205 ± 0.0006}    & \textcolor{dark2green}{0.7771 ± 0.0105} \\
HKdiagSE            & \textcolor{dark2green}{0.1202 ± 0.0004}    & \textcolor{dark2green}{0.7787 ± 0.0198} \\
CycleSE             & \textcolor{black}{0.1199 ± 0.0011}        & \textcolor{dark2green}{0.7739 ± 0.0240} \\
\bottomrule
\end{tabular}
% \vspace{-0.25in}
% \end{wraptable}
\end{table}

% \begin{table}[!htp]\centering
% \scriptsize
% \caption{GPSE training task ablation}\label{tab:abl_pt_task}
% \begin{tabular}{@{}cc|cc@{}}
% \toprule
% Excluded task & Finetuned? & \textbf{PCQM4Mv2} (subset) & \textbf{ogbg-molhiv} \\
% & &\textbf{MAE} $\downarrow$ &\textbf{AUROC} $\uparrow$ \\
% \midrule
% --                  & \xmark & 0.1196 ± 0.0004 & 0.7865 ± 0.0117 \\
% --                  & \cmark & 0.1195 ± 0.0007 & 0.7822 ± 0.0244 \\
% LapPE \& EigVals    & \xmark & 0.1200 ± 0.0006 & 0.7849 ± 0.0067 \\
% ElstaticPE          & \xmark & 0.1197 ± 0.0007 & 0.7681 ± 0.0146 \\
% RWSE                & \xmark & 0.1205 ± 0.0006 & 0.7771 ± 0.0105 \\
% HKdiagSE            & \xmark & 0.1202 ± 0.0004 & 0.7787 ± 0.0198 \\
% CycleSE             & \xmark & 0.1199 ± 0.0011 & 0.7739 ± 0.0240 \\
% \bottomrule
% \end{tabular}
% \end{table}

%% file: tables/abl_pt_dataset.tex
% \begin{table}[!htp]\centering
% \scriptsize
% \caption{GPSE training dataset ablation}\label{tab:abl_pt_ds}
% \begin{tabular}{@{}ccc@{}}
% \toprule
% % Dataset task & Finetuned? & \textbf{PCQM4Mv2} (subset) & \textbf{ogbg-molhiv} \todo{update} \\
% % & &\textbf{MAE} $\downarrow$ &\textbf{AUROC} $\uparrow$ \\
% &\multicolumn{2}{c}{\textbf{PCQM4Mv2} (subset)}\\
% Training dataset& Not finetuned & Finetuned \\
% \midrule
% ZINC        & 0.1202 ± 0.0005 & 0.1197 ± 0.0007 \\
% GEOM        & 0.1196 ± 0.0005 & 0.1194 ± 0.0002 \\
% ChEMBL      & 0.1195 ± 0.0003 & 0.1195 ± 0.0005 \\
% PCQM4Mv2    & 0.1192 ± 0.0005 & -- \\
% \midrule
% MolPCBA     & 0.1196 ± 0.0004 & 0.1195 ± 0.0007 \\
% \bottomrule
% \end{tabular}
% \end{table}

\begin{table}[!bp]\centering
% \scriptsize
% \fontsize{8.0pt}{8.0pt}\selectfont
\renewcommand{\arraystretch}{1.2}
\setlength\tabcolsep{6pt} % default value: 6pt
%\tiny
\caption{GPSE training dataset ablation. Performance measured in MAE $\downarrow$. \textcolor{dark2green}{Green} indicates fine-tuning \method on specific downstream dataset helps improve the performance. \textbf{Bold} indicates the best performance achieved on a particular downstream task.}\label{tab:abl_pt_ds}
\footnotesize
\begin{tabular}{@{}ccc|cc|cc@{}}
\toprule
& & &\multicolumn{2}{c}{\textbf{ZINC} (subset)}&\multicolumn{2}{c}{\textbf{PCQM4Mv2} (subset)}\\
Training dataset & \# unique graphs & Avg. \# nodes & Not finetuned & Finetuned & Not finetuned & Finetuned \\
\midrule
GEOM        & 169,925 & 18 & 0.0707 ± 0.0086 & \textcolor{dark2green}{0.0685 ± 0.0055} & 0.1196 ± 0.0005 & 0.1194 ± 0.0002 \\
ZINC        & 219,384 & 23 & 0.0700 ± 0.0041 & --              & 0.1202 ± 0.0005 & \textcolor{dark2green}{0.1197 ± 0.0007} \\
PCQM4Mv2    & 273,920 & 14 & 0.0721 ± 0.0042 & \textcolor{dark2green}{0.0713 ± 0.0014} & \textbf{0.1192 ± 0.0005} & --              \\
ChEMBL      & 970,963 & 30 & 0.0667 ± 0.0079 & \textbf{\textcolor{dark2green}{0.0643 ± 0.0036}} & 0.1195 ± 0.0003 & 0.1195 ± 0.0005 \\
\midrule
MolPCBA     & 323,555 & 25 & 0.0648 ± 0.0030 & 0.0668 ± 0.0076 & 0.1196 ± 0.0004 & 0.1195 ± 0.0007 \\
\bottomrule
\end{tabular}
\end{table}

%% file: tables/abl_gpse_lap.tex
\begin{table}[!htp]\centering
% \fontsize{8.0pt}{8.0pt}\selectfont
\caption{Ablation study on strategies to learn LapPE for \method.}
\footnotesize
\renewcommand{\arraystretch}{1.2}
\label{tab:abl_gpse_lap}
\vspace{3pt}
% \scriptsize
\begin{tabular}{@{}lcccc@{}}\toprule
& \textbf{ZINC (subset)} & \textbf{PCQM4Mv2 (subset)} & \textbf{Peptides-struct} & \textbf{Peptides-func} \\
& \textbf{MAE ↓} & \textbf{MAE ↓} & \textbf{MAE ↓} & \textbf{AP ↑} \\ \midrule
\method-abs (default) & \textbf{0.0957 ± 0.0044} & \textbf{0.1216 ± 0.0002} & \textbf{0.2516 ± 0.0018} & 0.6584 ± 0.0042 \\
\method-noabs & 0.1051 ± 0.0046 & 0.1229 ± 0.0006 & 0.2554 ± 0.0025 & \textbf{0.6687 ± 0.0119} \\
\method-signinvar & 0.1116 ± 0.0072 & 0.1243 ± 0.0004 & 0.2594 ± 0.0019 & 0.6619 ± 0.0097 \\
\method-SignNet & 0.1035 ± 0.0052 & 0.1232 ± 0.0006 & 0.2568 ± 0.0020 & 0.6647 ± 0.0093 \\
\bottomrule
\end{tabular}
\end{table}

%% file: tables/molnet_full.tex
\begin{sidewaystable}[!htp]\centering
\caption{Extended results for Table~\ref{tab:molnet}: Performance on MoleculeNet small datasets (scaffold test split), evaluated in AUROC (\%) $\uparrow$. \textcolor{dark2pink}{Red} indicates worse than baseline performance.}\label{tab:molnet_full}
\footnotesize
% \scriptsize
% \tiny
%\vspace{3pt}
\hspace*{-0.2cm}
\begin{tabular}{lccccccccc}\toprule
&\textbf{BBBP} &\textbf{BACE} &\textbf{Tox21} &\textbf{ToxCast} &\textbf{SIDER} &\textbf{ClinTox} &\textbf{MUV} &\textbf{HIV} \\ % &\textbf{Average} \\
\midrule
No pre-training (baseline)~\citep{Hu*2020Strategies} &65.8 ± 4.5 &70.1 ± 5.4 &74.0 ± 0.8 &63.4 ± 0.6 &57.3 ± 1.6 &58.0 ± 4.4 &71.8 ± 2.5 &75.3 ± 1.9 \\%& 66.96 \\
\midrule
SSL InfoMax pre-trained~\citep{Hu*2020Strategies} &68.8 ± 0.8 &75.9 ± 1.6 &75.3 ± 0.5 &\textcolor{dark2pink}{62.7 ± 0.4} &58.4 ± 0.8 &69.9 ± 0.3 &75.3 ± 2.5 &76.0 ± 0.7 \\ %& 70.29 \\
SSL EdgePred pre-trained~\citep{Hu*2020Strategies} &67.3 ± 2.4 &79.9 ± 0.9 &76.0 ± 0.6 &64.1 ± 0.6 &60.4 ± 0.7 &64.1 ± 3.7 &74.1 ± 2.1 &76.3 ± 1.0 \\ %&70.28 \\
SSL AttrMasking pre-trained~\citep{Hu*2020Strategies} &\textcolor{dark2pink}{64.3 ± 2.8} &79.3 ± 1.6 &76.7 ± 0.4 &64.2 ± 0.5 &61.0 ± 0.7 &71.8 ± 4.1 &74.7 ± 1.4 &77.2 ± 1.1 \\ %&71.15 \\
SSL ContextPred pre-trained~\citep{Hu*2020Strategies} &68.0 ± 2.0 &79.6 ± 1.2 &75.7 ± 0.7 &63.9 ± 0.6 &60.9 ± 0.6 &65.9 ± 3.8 &75.8 ± 1.7 &77.3 ± 1.0 \\ %&70.89 \\
GraphCL pre-trained~\citep{you2020graph} &69.7 ± 0.7 &75.4 ± 1.4 &\textcolor{dark2pink}{73.9 ± 0.7} &\textcolor{dark2pink}{62.4 ± 0.6} &60.5 ± 0.9 &76.0 ± 2.7 &\textcolor{dark2pink}{69.8 ± 2.7} &\textbf{78.5 ± 1.2} \\ %&70.78 \\
InfoGraph pre-trained~\citep{wang2022evaluating} &66.3 ± 0.6 &\textcolor{dark2pink}{64.8 ± 0.8} &\textcolor{dark2pink}{68.1 ± 0.6} &\textcolor{dark2pink}{58.4 ± 0.6} &\textcolor{dark2pink}{57.1 ± 0.8} &66.3 ± 0.6 &\textcolor{dark2pink}{44.3 ± 0.6} &\textcolor{dark2pink}{70.2 ± 0.6}& \\ %\textcolor{dark2pink}{61.94} \\
JOAOv2 pre-trained~\citep{wang2022evaluating} &66.4 ± 0.9 &\textcolor{dark2pink}{67.4 ± 0.7} &\textcolor{dark2pink}{68.2 ± 0.8} &\textcolor{dark2pink}{57.0 ± 0.5} &59.1 ± 0.7 &64.5 ± 0.9 &\textcolor{dark2pink}{47.4 ± 0.8} &\textcolor{dark2pink}{68.4 ± 0.5} \\ %&\textcolor{dark2pink}{62.30} \\
GraphMAE~\citep{hou2022graphmae} &72.0 ± 0.6 &83.1 ± 0.9 &75.5 ± 0.6 &64.1 ± 0.3 &60.3 ± 1.1 &\textbf{82.3 ± 1.2} &76.3 ± 2.4 &77.2 ± 1.0 \\ %& 73.85 \\
GraphLoG~\citep{xu2021self} &\textbf{72.5 ± 0.8} &\textbf{83.5 ± 1.2} &75.7 ± 0.5 &63.5 ± 0.7 &61.2 ± 1.1 &76.7 ± 3.3 & 76.0 ± 1.1 & 77.8 ± 0.8 \\ %& 73.36 \\
\midrule
GraphLoG augmented &\textcolor{dark2pink}{65.6 ± 1.0} &82.5 ± 1.2 &\textcolor{dark2pink}{73.2 ± 0.5} &63.6 ± 0.4 &60.9 ± 0.7 &72.5 ± 3.5 &72.4 ± 1.5 &\textcolor{dark2pink}{74.4 ± 1.5} \\
LapPE augmented &67.1 ± 1.6 &80.4 ± 1.5 &76.6 ± 0.3 &65.9 ± 0.7 &59.3 ± 1.7 &76.4 ± 2.3 &75.6 ± 0.8 &75.6 ± 1.1 \\ %&71.63 \\
RWSE augmented &67.0 ± 1.4 &79.6 ± 2.8 &76.3 ± 0.5 &65.6 ± 0.3 &58.5 ± 1.4 &74.5 ± 4.4 &75.0 + 1.0 &78.1 ± 1.5 \\ %&71.39 \\
AllPSE augmented &67.6 ± 1.2 &77.0 ± 4.4 &75.9 ± 1.0 &63.9 ± 0.3 &\textbf{63.0 ± 0.6} &72.6 ± 4.3 &\textcolor{dark2pink}{67.9 ± 0.7} &75.4 ± 1.5 \\ %&70.41 \\
\textbf{\method} augmented &66.2 ± 0.9 &80.8 ± 3.1 &\textbf{77.4 ± 0.8} &\textbf{66.3 ± 0.8} &61.1 ± 1.6 &78.8 ± 3.8 &\textbf{76.6 ± 1.2} &77.2 ± 1.5 \\ %&73.29 \\
%GPSE augmented (MNIST) &67.5 ± 1.3 &79.0 ± 1.4 &76.2 ± 0.7 &\textbf{66.0 ± 0.6} &61.1 ± 0.6\\

\bottomrule
\end{tabular}
\end{sidewaystable}